\documentclass{article}

\usepackage{arxiv}

\usepackage[utf8]{inputenc} % allow utf-8 input
\usepackage[T1]{fontenc}    % use 8-bit T1 fonts
\usepackage{hyperref}       % hyperlinks
\usepackage{url}            % simple URL typesetting
\usepackage{booktabs}       % professional-quality tables
\usepackage{amsfonts}       % blackboard math symbols
\usepackage{amsmath}
\usepackage{nicefrac}       % compact symbols for 1/2, etc.
\usepackage{microtype}      % microtypography
\usepackage{lipsum}		% Can be removed after putting your text content
\usepackage{graphicx}
\usepackage{natbib}
\usepackage{doi}
\usepackage{tcolorbox}
\usepackage{amsmath}
\usepackage{amssymb}
\usepackage{algorithm}
\usepackage{algpseudocode}
\usepackage{caption}
\usepackage{placeins}

\usepackage{multirow}

\title{CTRL-FM: Inference-Time Structural Control for Flow-Based Virtual Sensing under Long-Horizon Node Outages}

\author{ {\hspace{1mm}Boshra Khalili} \\
	Department of Civil Engineering \\and Engineering Mechanics\\
    Columbia University\\
	New York, NY 10027, USA \\
	\texttt{bk2898@columbia.edu} \\
    \And
    {\hspace{1mm}Andrew Smyth} \\
    	Department of Civil Engineering\\ and Engineering Mechanics\\
        Columbia University\\
    	New York, NY 10027, USA \\
    	\texttt{aws16@columbia.edu} \\
    }
\renewcommand{\shorttitle}{CTRL-FM}

\hypersetup{
pdftitle={CTRL-FM: Inference-Time Structural Control for Flow-Based Virtual Sensing under Long-Horizon Node Outages},
pdfsubject={Intelligent transportation system},
pdfauthor={Boshra Khalili, Andrew Smyth},
pdfkeywords={Spatiotemporal Graph Imputation · Flow Matching · Inference-Time Structural Control · Long-Horizon Sensor Outages · Self-Supervised Policy Selection},
}

\begin{document}
\maketitle

\begin{abstract}
Long-horizon node outages are harder than conventional missing-value imputation because complete sensor trajectories remain unobserved for extended periods, while the usefulness of spatial and temporal dependencies varies across outage locations and operating regimes. Existing graph, diffusion, and flow-based methods typically impose a fixed structural bias, which can oversmooth abrupt changes or propagate errors from unreliable neighbors. We introduce CTRL-FM, an observation-preserving flow-matching framework that treats structural guidance as an outage-conditioned inference-time decision. A missing-only flow-matching backbone learns a conditional velocity field while keeping observed measurements exactly fixed. For each realized outage, CTRL-FM constructs self-supervised proxy outages from surviving measurements and uses their known values to select among uncontrolled, graph-dominant, temporal-dominant, and balanced control policies. The selected policy is locally modulated by graph and temporal reliability estimated exclusively from surviving observations, while boundary-anchored trends provide structural guidance. Control is applied only to missing entries and is rejected when proxy evidence does not sufficiently improve over uncontrolled sampling. Across PeMS08, METR-LA, Air-36, and AQI, CTRL-FM achieves the strongest overall performance under 30\% and 60\% node outages. Stress tests with up to 75\% sensor loss further show graceful degradation and consistent gains over training-time and fixed structural control. These results demonstrate that validating, selecting, and locally adapting structural guidance for each realized outage is more effective than imposing a single global structural prior.
\end{abstract}


\section{Introduction}
\label{sec:introduction}

Spatio-temporal graph imputation, also known as graph kriging or virtual
sensing, reconstructs signals at unavailable locations from measurements
observed across time and a sensor network. It supports applications in
traffic monitoring, environmental sensing, climate analysis, and urban
infrastructure management~\cite{
Yang2025GraphStructureLearningImputation,
Chen2025FourthOrderTensorCompletion,
Ye2021GATTrafficImputation,
Shen2023BidirectionalGATRecurrentImputation}.
Given a multivariate graph signal, the goal is to recover missing node
trajectories by exploiting temporal dynamics and dependencies among connected
locations.

Most existing methods focus on randomly missing entries, short gaps, or
moderate sparsity. Real sensing systems also experience long-horizon node
outages, in which complete sensors remain unavailable over extended
contiguous intervals because of hardware failure, maintenance, communication
loss, or sparse deployment~\cite{
Zhong2023IterativeGraphReconstructionImputation,
Jia2025ReviewGNNProcessMonitoring,
He2025SpatiotemporalGCNWaterLevel}.
This setting is substantially harder: failed nodes provide no direct evidence
during the outage, temporal anchors may be distant, and multiple trajectories
may be compatible with the surviving observations.

Graph neural networks, diffusion models, and flow-based methods reduce this
ambiguity by imposing graph connectivity, temporal continuity, or signal
smoothness~\cite{
wu2021ignnk,
cini2021grin,
Zhang2025Diffputer,
giflow2025}.
However, the reliability of these assumptions depends on the realized outage
and operating regime. Neighboring traffic sensors may enter different
congestion states, environmental signals may change abruptly, and graph
relations may become misleading during incidents, drift, or operational
interventions~\cite{
Cuza2022StochasticTrafficSpeedImputation,
Xu2024HierarchicalSTGCNImputation, Khalili2026STREAMOOD}.
Consequently, a structural prior that improves one outage may oversmooth
another or propagate errors from unrepresentative neighbors.

Existing approaches typically encode structural guidance as a global modeling
choice: it is embedded in the architecture, learned through training-time
regularization, or imposed through a fixed sampling objective. Such strategies
identify what is useful on average over historical masks, but they do not
determine whether the same graph--temporal assumption is appropriate for the
specific outage encountered at inference. Because the failed values are
unknown, the model also lacks a direct mechanism for validating or rejecting
its structural choice.

We address this limitation with CTRL-FM, an
observation-preserving flow-matching framework that treats structural guidance
as an outage-conditioned inference-time control decision. CTRL-FM separates
general reconstruction learning from outage-specific structural control. A
missing-only flow-matching backbone learns a reusable conditional velocity
field while keeping observed measurements fixed throughout training and
sampling. The backbone is then frozen, and its sampling velocity is adapted
for each realized outage rather than retraining the model or applying a
post-processing correction.

At inference, CTRL-FM temporarily hides nearby surviving measurements to
construct local self-supervised proxy outages. Because their values remain
available as evaluation targets, these proxies allow the method to compare
uncontrolled, graph-dominant, temporal-dominant, and balanced control policies
under the current outage context. The selected policy is applied to the true
outage only when it sufficiently improves over uncontrolled sampling;
otherwise, CTRL-FM abstains from structural intervention.

CTRL-FM further separates which policy should control an outage from
where that policy should be trusted. Reliability modules
estimate graph and temporal support exclusively from surviving observations,
using local coverage, neighborhood agreement, change magnitude, graph
distance, and proximity to observed temporal boundaries. Boundary-anchored
guidance then connects the evolving flow velocity to valid observed trends,
while missing-region projection and repeated clamping preserve every observed
measurement exactly.

We evaluate CTRL-FM on the PeMS08 and METR-LA traffic networks and the Air-36
and AQI air-quality networks under 30\% and 60\% long-horizon node outages,
with additional stress tests up to 75\% sensor loss. Beyond reconstruction
accuracy, we evaluate whether proxy outages provide a reliable
self-supervised basis for policy selection through proxy coverage, agreement
with retrospective oracle rankings, and oracle-policy match.
Our contributions are:
\begin{itemize}
    \item We formulate structural guidance as an outage-conditioned
    inference-time control problem and use local self-supervised proxy
    outages to select how the frozen flow velocity should be modified for
    each realized failure.

    \item We combine observed-only reliability, boundary-anchored guidance,
    and conservative abstention to determine where structural correction is
    supported and when it should be rejected.

    \item We develop an observation-preserving, missing-only flow-matching
    formulation that confines transport and structural control to unavailable
    entries while keeping all surviving measurements exactly fixed.
\end{itemize}

The remainder of the paper is organized as follows. The Method section
presents CTRL-FM, while the Experiments section evaluates reconstruction
accuracy, efficiency, proxy validity, and robustness under severe outages.

\section{Related Work}
\label{sec:related_work}

Classical spatio-temporal imputation methods rely on low-rank, tensor,
statistical, or kernel assumptions~\cite{
Zhou2015SpatioTemporalTensorCompletion,
Lei2022BayesianKernelizedMFImputation,
Wu2021SpatialAggregationTemporalConvKriging}.
Neural approaches learn richer dependencies directly from data.
BRITS~\cite{cao2018brits} and SAITS~\cite{du2023saits} primarily model
temporal and cross-variable relationships, whereas
IGNNK~\cite{wu2021ignnk}, KCN~\cite{appleby2020kcn},
GRIN~\cite{cini2021grin}, SPIN~\cite{marisca2022spin}, and
OPCR~\cite{liu2024opcr} exploit graph structure for spatial reconstruction.
Inductive and graph-learning methods additionally support inference at
partially or entirely unavailable locations~\cite{
wu2021ignnk,
appleby2020kcn,
Zhong2023IterativeGraphReconstructionImputation,
Yang2025GraphStructureLearningImputation}.
However, these methods generally apply a single learned reconstruction
strategy across all outage configurations.

Probabilistic methods address the ambiguity of missing trajectories by
modeling conditional distributions. CSDI~\cite{tashiro2021csdi} introduces
score-based diffusion for time-series imputation, while
PriSTI~\cite{liu2023pristi}, CDSTE~\cite{lei2024cdste},
Diffputer~\cite{Zhang2025Diffputer}, and
SSD-TS~\cite{Gao2025SSDTS} incorporate temporal, spatial, or geographic
structure into diffusion-based reconstruction. Their structural conditioning,
however, is typically fixed during training and reused for every test outage.

Flow matching provides a continuous-time alternative to iterative diffusion
sampling~\cite{Lipman2022FlowMatching,Zhang2025HierarchicalRectifiedFlow}.
GiFlow~\cite{giflow2025} applies graph-informed flow matching to
spatio-temporal imputation, while training-free guidance methods modify frozen
flow models through auxiliary sampling objectives~\cite{
Wang2024TrainingFreeGuidedFlowMatching,
Xu2025EnergyGuidedFlowMatching,
WanElucidatingFlowMatchingDynamics}.
These methods commonly use a predefined structural objective or guidance
strength, which may be unreliable when graph and temporal dependencies vary
across outages.

CTRL-FM instead treats structural guidance as an outage-conditioned
inference-time decision. It uses self-supervised proxy outages to select among
graph, temporal, balanced, and uncontrolled policies, modulates the selected
policy using reliability estimated only from surviving observations, and
abstains when structural control is not supported. Thus, CTRL-FM validates
both when and where structural guidance should be applied for each realized
outage.

\section{Method}
\label{sec:method}

\subsection{Outage-Conditioned Controlled Flow Matching}
\label{sec:missing_only_fm}

Conventional conditional flow matching~\cite{Lipman2022FlowMatching} learns a velocity field whose behavior
is fixed after training. Although the field may condition on observations,
masks, and graph structure, the same learned structural bias is reused across
outages with different failed nodes and operating regimes. This is restrictive
for long-horizon failures: graph neighbors may be informative for one outage
but misleading for another, while strong temporal continuity may suppress
peaks, drops, or regime transitions.

CTRL-FM separates general reconstruction learning from outage-specific
structural control. A missing-only flow-matching backbone learns an
observation-preserving transport field. At inference, local self-supervised
proxy outages determine how the frozen field should be controlled, while
observed-only reliability determines where the selected graph and temporal
guidance should be trusted. The resulting correction is injected directly
into the sampling velocity rather than applied after reconstruction.
\begin{figure*}
    \centering
    \includegraphics[width=1\linewidth]{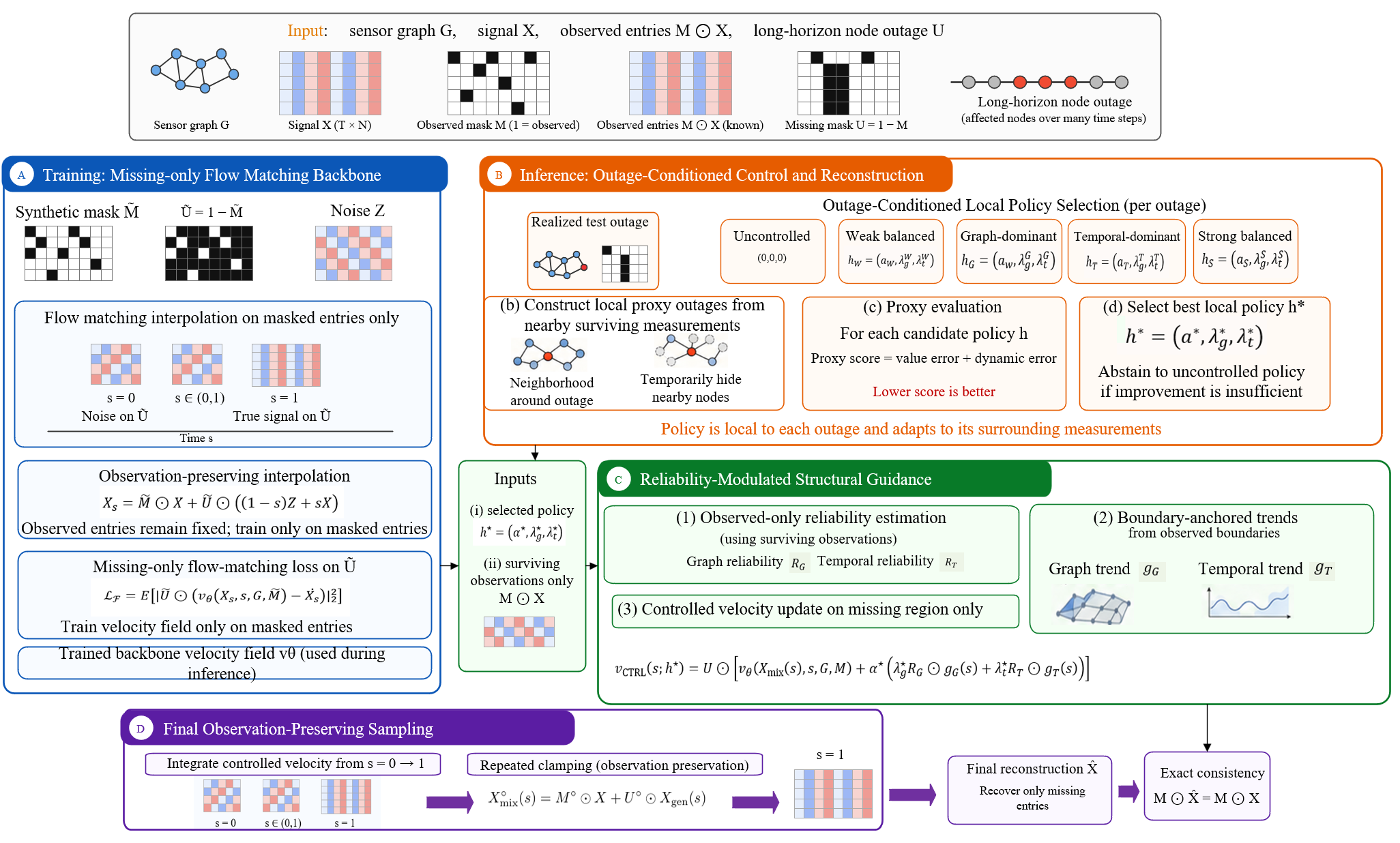}
    \caption{Overview of CTRL-FM. A missing-only flow backbone is trained while
    preserving observations. At inference, local proxy outages select an
    outage-specific policy, observed-only reliability modulates structural
    guidance, and the controlled velocity reconstructs only missing entries
    while maintaining exact observation consistency.}
    \label{fig:placeholder}
\end{figure*}

\paragraph{Problem setting.}
Let \(G=(V,E)\) be a sensor graph with \(K=|V|\) nodes, and let
\(X\in\mathbb{R}^{T\times K\times C}\) denote a \(C\)-channel
spatio-temporal signal. The observation and missing masks are
\begin{equation}
M\in\{0,1\}^{T\times K\times C},
\qquad
U=1-M,
\label{eq:problem_masks}
\end{equation}
where \(M_{t,k,c}=1\) indicates an observed entry. A long-horizon node outage
removes all channels of nodes \(\mathcal O\subseteq V\) over a contiguous
interval \(\mathcal I\):
\begin{equation}
M_{t,k,c}=0,
\qquad
k\in\mathcal O,\;
t\in\mathcal I,\;
c=1,\ldots,C.
\label{eq:long_outage}
\end{equation}
The objective is to reconstruct the unavailable trajectories while preserving
all surviving measurements exactly:
\begin{equation}
M\odot\widehat X=M\odot X.
\label{eq:exact_consistency}
\end{equation}

\paragraph{Observation-preserving missing-only flow matching.}
During training, CTRL-FM samples a synthetic outage mask
\(\widetilde M\), with \(\widetilde U=1-\widetilde M\). Given Gaussian noise
\(Z\sim\mathcal N(0,I)\) and flow time \(s\sim\mathcal U(0,1)\), the
conditional transport path is
\begin{equation}
X_s
=
\widetilde M\odot X
+
\widetilde U\odot
\bigl((1-s)Z+sX\bigr).
\label{eq:missing_only_path}
\end{equation}
Noise and transport are restricted to the synthetically missing region.
Therefore,
\begin{equation}
\widetilde M\odot X_s
=
\widetilde M\odot X,
\qquad
s\in[0,1],
\label{eq:training_preservation}
\end{equation}
so all observed measurements remain fixed throughout the training path.

The target velocity and training objective are
\begin{equation}
\begin{aligned}
v^\star
&=
\widetilde U\odot(X-Z),
\\
\mathcal L_{\mathrm{FM}}
&=
\mathbb E
\left[
\frac{
\left\|
\widetilde U\odot
\left(
v_\theta(X_s,s,G,\widetilde M)-v^\star
\right)
\right\|_F^2
}{
\|\widetilde U\|_1+\epsilon_{\mathrm{mask}}
}
\right].
\end{aligned}
\label{eq:missing_only_training}
\end{equation}
Thus, the backbone is optimized only where reconstruction is required and
learns a reusable conditional velocity field without embedding one fixed
graph--temporal control policy into its parameters.

\paragraph{Outage-conditioned control policies.}
At inference, CTRL-FM considers a compact policy library
\begin{equation}
\mathcal H
=
\{h_0,h_1,\ldots,h_R\},
\qquad
h_r
=
(\alpha_r,\lambda_{g,r},\lambda_{t,r}),
\qquad
h_0=(0,0,0),
\label{eq:candidate_policies}
\end{equation}
where $\alpha_r$ controls the total structural-correction strength, while
$\lambda_{g,r}$ and $\lambda_{t,r}$ determine its graph--temporal balance.
These policies are not separately trained models; they define alternative
ways of controlling the same frozen flow-matching backbone. The complete candidate library is provided in Appendix Section 1.2, with proxy calibration and validation settings in Appendix Sections 1.3 and 2.3.

\paragraph{Self-supervised local proxy outages.}
Because the true failed values are unavailable, CTRL-FM constructs proxy
outages from nearby surviving nodes:
\begin{equation}
\mathcal V_{\mathrm{proxy}}
=
\left\{
k\in V\setminus\mathcal O:
d_G(k,\mathcal O)\leq r_p
\right\},
\label{eq:proxy_neighborhood}
\end{equation}
where $d_G$ denotes graph distance and $r_p$ is the proxy radius. For proxy
$b$, a mask $P^{(b)}$ temporarily hides originally observed measurements:
\begin{equation}
P^{(b)}\odot U=0,
\qquad
M^{(b)}
=
M\odot\left(1-P^{(b)}\right).
\label{eq:proxy_masking}
\end{equation}
The first condition prevents the proxy from overlapping the true outage,
which remains active throughout calibration. The temporarily hidden values
are retained only as self-supervised evaluation targets.

For each proxy $b$ and policy $h_r$, CTRL-FM performs a short controlled-flow
probe:
\begin{equation}
\widehat X^{(b,r)}
=
\operatorname{ProbeSample}
\left(
M^{(b)}\odot X,\,
M^{(b)},\,
G,\,
h_r
\right).
\label{eq:proxy_reconstruction}
\end{equation}
The backbone input, boundary anchors, structural guidance, and reliability
features are all recomputed under $M^{(b)}$. Thus, temporarily hidden proxy
values cannot contribute to their own reconstruction or reliability
estimation and are accessed only afterward to score the candidate policy.

The proxy objective combines pointwise reconstruction accuracy and
temporal-change recovery:
\begin{equation}
\begin{aligned}
\ell_{b,r}
={}&
\frac{
\left\|
P^{(b)}\odot
\left(
\widehat X^{(b,r)}-X
\right)
\right\|_1
}{
\|P^{(b)}\|_1+\epsilon_p
}
\\[-1mm]
&+
\lambda_{\mathrm{dyn}}
\frac{
\left\|
Q^{(b)}\odot
\left(
\Delta_t\widehat X^{(b,r)}
-
\Delta_tX
\right)
\right\|_1
}{
\|Q^{(b)}\|_1+\epsilon_p
},
\end{aligned}
\label{eq:proxy_loss}
\end{equation}
where
$(\Delta_tY)_{t,k,c}=Y_{t,k,c}-Y_{t-1,k,c}$ and $Q^{(b)}$ retains valid
temporal differences involving proxy entries. The dynamic term discourages
policies that achieve low pointwise error by oversmoothing abrupt changes.
Proxy eligibility, interval selection, and the construction of $Q^{(b)}$ are
detailed in Appendix Section 1.3.

\paragraph{Local policy selection and conservative abstention.}
Proxies closer to the true failed region receive larger weights:
\begin{equation}
\omega_b
=
\frac{
\exp(-d_b/\tau_p)
}{
\sum_{b'\in\mathcal B}
\exp(-d_{b'}/\tau_p)
},
\qquad
\mathcal J_{\mathrm{proxy}}(h_r)
=
\sum_{b\in\mathcal B}
\omega_b\,\ell_{b,r},
\label{eq:proxy_policy_score}
\end{equation}
where $\mathcal B$ is the set of valid proxies and $d_b$ is the mean graph
distance between proxy $b$ and the true failed-node set. The locally preferred
policy is
\begin{equation}
\widetilde h
=
\arg\min_{h_r\in\mathcal H}
\mathcal J_{\mathrm{proxy}}(h_r).
\label{eq:local_policy_selection}
\end{equation}
CTRL-FM accepts the preferred policy only when its relative improvement over
uncontrolled sampling exceeds the abstention margin:
\begin{equation}
h^\star
=
\begin{cases}
\widetilde h,
&
\dfrac{
\mathcal J_{\mathrm{proxy}}(h_0)
-
\mathcal J_{\mathrm{proxy}}(\widetilde h)
}{
\mathcal J_{\mathrm{proxy}}(h_0)
+
\epsilon_{\mathrm{abst}}
}
\geq
\delta_{\mathrm{abst}},
\\[11pt]
h_0,
& \text{otherwise}.
\end{cases}
\label{eq:final_policy}
\end{equation}
If no valid proxy can be constructed, CTRL-FM also defaults to $h_0$.
Structural intervention is therefore applied only when supported by local
self-supervised evidence. Complete weighting, tie-breaking, and fallback
details are provided in Appendix Section 1.3.

\paragraph{Observation-preserving active state.}
Let $M^\circ=M^{(b)}$ during proxy calibration and $M^\circ=M$ during final
inference, with $U^\circ=1-M^\circ$. At flow time $s$, the frozen backbone
receives
\begin{equation}
X_{\mathrm{mix}}^\circ(s)
=
M^\circ\odot X
+
U^\circ\odot X_{\mathrm{gen}}(s),
\label{eq:inference_mixed_state}
\end{equation}
which fixes all currently observed entries and evolves only unavailable ones.

\paragraph{Boundary-anchored structural guidance.}
CTRL-FM uses valid physical-time changes from surviving measurements to
guide the unavailable region. These changes provide observed-side anchors
where consecutive measurements are available. Let \(d_{\mathrm{obs}}\)
denote the valid observed changes and \(Q\) their support mask. We define
\begin{equation}
A(s)
=
\operatorname{sg}\!\left[
U^\circ \odot
v_\theta\!\left(X_{\mathrm{mix}}^\circ(s),s,G,M^\circ\right)
\right]
+
M^\circ \odot Q \odot \kappa d_{\mathrm{obs}},
\label{eq:comparison_field}
\end{equation}
where \(\kappa\) aligns the observed-change and flow-velocity scales, and
\(\operatorname{sg}\) denotes stop-gradient. The first term contains the
backbone velocity on unavailable entries, while the second inserts
measurement-supported trends for structural comparison only; observed
measurements are never modified.

Graph and temporal compatibility are measured by
\begin{equation}
\begin{aligned}
C_G(A)
&=
\operatorname{Avg}_{t,(i,j)\in E}
\left[
w_{ij}\,d_\rho(A_{t,i,:},A_{t,j,:})
\right],
\\
C_T(A)
&=
\operatorname{Avg}_{t,k}
\left[
d_\rho(A_{t+1,k,:},A_{t,k,:})
\right].
\end{aligned}
\label{eq:compatibility_objectives}
\end{equation}
The corresponding guidance directions are
\begin{equation}
g_G(s)=-\nabla_A C_G(A(s)),
\qquad
g_T(s)=-\nabla_A C_T(A(s)).
\label{eq:structural_guidance}
\end{equation}
They encourage spatial and temporal compatibility of the missing-region
velocity. Because the backbone term is stop-gradient, no differentiation
through \(v_\theta\) is required. Complete definitions of
\(d_{\mathrm{obs}}\), \(Q\), \(d_\rho\), and \(w_{ij}\) are provided in
Appendix Sections~1.4--1.5.

\paragraph{Observed-only reliability.}
The selected policy determines the overall graph--temporal control, while
reliability determines where each component should be trusted. Let
$X_{\mathrm{obs}}^\circ=M^\circ\odot X$. CTRL-FM computes
\begin{equation}
\begin{aligned}
R_G
&=
S_G\odot
\sigma
\left(
f_g
\left(
\phi_G(X_{\mathrm{obs}}^\circ,M^\circ)
\right)
\right),
\\
R_T
&=
S_T\odot
\sigma
\left(
f_t
\left(
\phi_T(X_{\mathrm{obs}}^\circ,M^\circ)
\right)
\right),
\end{aligned}
\label{eq:observed_only_reliability}
\end{equation}
where $\phi_G$ and $\phi_T$ summarize observed support, neighborhood
agreement, change magnitude, graph distance, and proximity to an observed
temporal boundary. The networks $f_g$ and $f_t$ are frozen reliability
modules, while $S_G$ and $S_T$ gate reliability to zero where no valid
observational support is available.

All features, support maps, and reliability values are recomputed under the
active mask. Temporarily hidden proxy values therefore cannot influence
their own reliability estimates. The reliability modules are trained
self-supervisedly using synthetic training outages and are frozen before
test-time calibration. Detailed feature and training definitions are
provided in Appendix Section 1.6.

\paragraph{Outage-conditioned controlled velocity.}
For the selected policy
\[
h^\star
=
(\alpha^\star,\lambda_g^\star,\lambda_t^\star),
\]
CTRL-FM defines the final sampling dynamics as
\begin{equation}
\begin{aligned}
\frac{\mathrm dX_{\mathrm{gen}}(s)}{\mathrm ds}
&=
v_{\mathrm{CTRL}}(s;h^\star)
\\
&=
U\odot
\Big[
v_\theta
\left(
X_{\mathrm{mix}}(s),s,G,M
\right)
\\[-1mm]
&\hspace{14mm}
+
\alpha^\star
\left(
\lambda_g^\star R_G\odot g_G(s)
+
\lambda_t^\star R_T\odot g_T(s)
\right)
\Big].
\end{aligned}
\label{eq:controlled_velocity}
\end{equation}
The policy $h^\star$ determines the outage-level correction strength and
graph--temporal balance, whereas $R_G$ and $R_T$ determine where the
corresponding guidance fields are credible. Projection by $U$ confines both
the backbone velocity and every structural correction to missing entries.

Equation~\eqref{eq:controlled_velocity} is the central algorithmic extension
of CTRL-FM. Rather than integrating the same learned or externally guided
velocity for every outage, CTRL-FM constructs an outage-specific velocity by
selecting, reliability-modulating, and injecting structural control during
sampling. When $h^\star=h_0$, the correction vanishes and the method reduces
exactly to the uncontrolled missing-only flow-matching backbone.

\paragraph{Observation-preserving integration.}
After every numerical solver stage, CTRL-FM re-clamps the state using
Eq.~\eqref{eq:inference_mixed_state}. Consequently,
\begin{equation}
M\odot X_{\mathrm{mix}}(s)
=
M\odot X,
\qquad
s\in[0,1],
\label{eq:exact_preservation_inference}
\end{equation}
so neither the backbone velocity nor the structural controller can alter an
observed measurement. The final reconstruction is
\begin{equation}
\widehat X
=
M\odot X
+
U\odot X_{\mathrm{gen}}(1).
\label{eq:final_imputation}
\end{equation}
Thus, one frozen flow-matching backbone can realize outage-specific transport
dynamics without retraining, test-time parameter updates, or access to true
failed values. The formal invariance argument and complete calibration and
sampling algorithms are provided in Appendix Sections 1.7--1.8.

% ===========================================================

\section{Experiments}
\label{sec:experiments}

\subsection{Experimental Setup}
\label{sec:experimental_setup}
\begin{figure*}[t]
    \centering
    \includegraphics[width=1\linewidth]{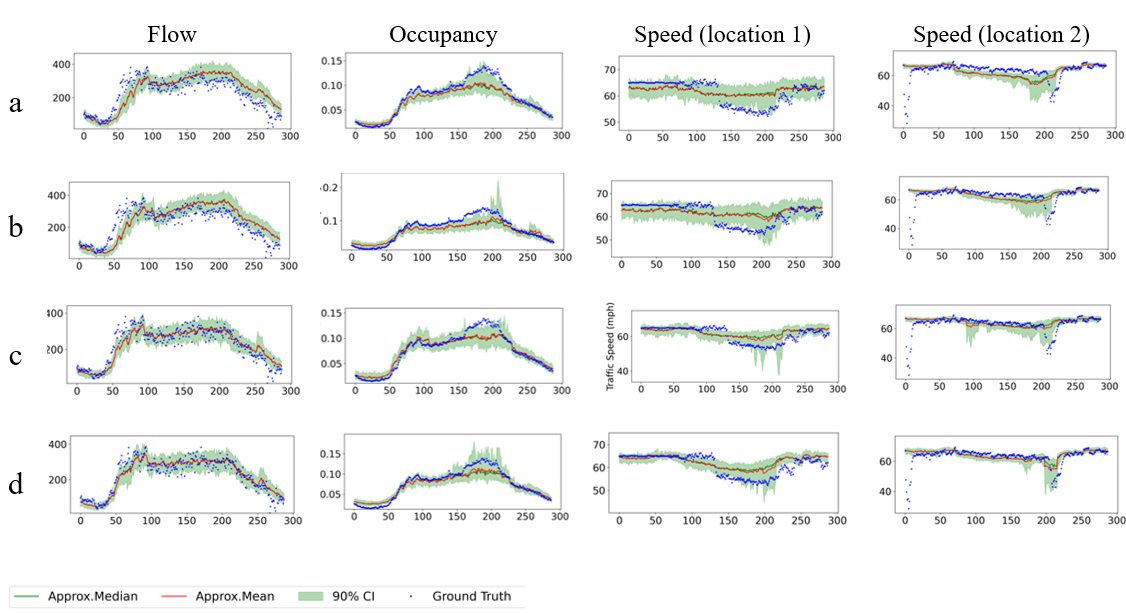}
    \caption{
    Qualitative PeMS08 reconstruction under a 60\% long-horizon node outage.
    Rows (a)--(d) show CSDI, FM-TrainStruct, CDSTE, and CTRL-FM,
    respectively; columns show flow, occupancy, and speed at two failed
    sensors. Blue points denote ground truth, red and green curves denote
    the predictive mean and median, and shading denotes the 90\% predictive
    interval.
    }
    \label{fig:qualitative_node_outage}
\end{figure*}
\paragraph{Datasets and outage protocol.}
We evaluate CTRL-FM on two traffic datasets,
PeMS08~\cite{Chen2001PeMS,Guo2019ASTGCN} and
METR-LA~\cite{Li2018DCRNN}, and two air-quality datasets,
Air-36~\cite{Yi2016STMVL} and AQI~\cite{Zheng2015AirQuality}.
Long-horizon outages are simulated by removing the complete trajectories of
randomly selected nodes at outage ratios of 30\% and 60\%. For multivariate
PeMS08, the same node mask is applied to flow, occupancy, and speed. All
methods use identical chronological splits, outage masks, test windows, and
failed-region evaluation sets. Dataset statistics, preprocessing, graph
construction, and the complete outage protocol are provided in Appendix
Section 2.1.

\paragraph{Baselines and controlled variants.}
We compare CTRL-FM with a non-learning Graph Neighbor baseline;
deterministic neural imputers
SAITS~\cite{du2023saits},
IGNNK~\cite{wu2021ignnk},
GRIN~\cite{cini2021grin}, and
OPCR~\cite{liu2024opcr};
probabilistic models
CSDI~\cite{tashiro2021csdi},
PriSTI~\cite{liu2023pristi}, and
CDSTE~\cite{lei2024cdste};
and the graph-informed flow-matching method
GiFlow~\cite{giflow2025}.
CDSTE is designed and evaluated only for traffic datasets. FM-TrainStruct uses
the same backbone as CTRL-FM but incorporates structural regularization only
during training. Fixed-CTRL applies one validation-selected control policy to
all test outages, whereas CTRL-FM selects and modulates the policy separately
for each realized outage. Detailed baseline descriptions are provided in
Appendix Section 2.4.

\paragraph{Implementation and evaluation.}
Hyperparameters are selected on the validation split without access to test-outage values, and results are averaged over three independent runs. MAE measures pointwise error, CRPS~\cite{gneiting2007strictly} evaluates probabilistic quality, and Peak-$\Delta$MAE evaluates large temporal changes. MAE and Peak-$\Delta$MAE are reported in original physical units after inverse normalization, whereas CRPS is computed in standardized space. All metrics are evaluated only on the true failed-node region; lower values are better. Training, calibration, and metric details are provided in Appendix Sections 2.2, 2.3, and 2.5.

% ============================================================
\subsection{Validity of Proxy-Based Policy Selection}
\label{sec:proxy_validity}

This experiment tests whether nearby observed measurements can reliably guide policy selection. Valid-proxy coverage shows how often enough proxy data are available. Spearman’s \(\rho\)\ measures whether proxy scores rank policies similarly to their true missing-region errors. Oracle match shows how often CTRL-FM selects the same best policy as a post-hoc oracle, which determines the best policy after reconstruction by comparing all policies against the true missing values.
The post-hoc oracle is used only for retrospective diagnosis after all
policy-selection, abstention, and reconstruction decisions have been
completed. It is never used for proxy construction, policy scoring,
hyperparameter selection, or final reconstruction.

\begin{table}[t]
\centering
\scriptsize
\setlength{\tabcolsep}{4.5pt}
\renewcommand{\arraystretch}{1.06}
\caption{
Validity of proxy-based policy selection at 60\% node outage. Valid proxy
measures calibration coverage, Spearman's \(\rho\) measures agreement with
the post-hoc true policy ranking, and oracle match reports selection of the
true best policy. Higher values are better.
}
\label{tab:main_proxy_validity}
\resizebox{\linewidth}{!}{%
\begin{tabular}{lccc}
\toprule
\textbf{Dataset}
& \textbf{Valid proxy (\%)$\uparrow$}
& $\boldsymbol{\rho}\uparrow$
& \textbf{Oracle match (\%)$\uparrow$} \\
\midrule
PeMS08  & 98.7 & \textbf{0.82} & \textbf{88.4} \\
METR-LA & \textbf{99.1} & 0.79 & 84.8 \\
Air-36  & 87.4 & 0.68 & 78.6 \\
AQI     & 98.9 & 0.76 & 81.6 \\
\bottomrule
\end{tabular}%
}
\end{table}

As shown in Table~\ref{tab:main_proxy_validity}, valid proxies are available
for 98.7--99.1\% of outages on PeMS08, METR-LA, and AQI. Coverage decreases
to 87.4\% on Air-36 because its network provides fewer surviving
proxy candidates. Proxy rankings achieve correlations of 0.68--0.82 with
the post-hoc oracle, and CTRL-FM selects the true best candidate in
78.6--88.4\% of valid cases. These results support the use of local
surviving measurements for policy selection without revealing failed-node
values. Formal definitions are provided in Appendix Section 2.5, and
selected-policy frequencies are reported in Appendix Section 2.8.

% ============================================================
\subsection{Reconstruction under Long-Horizon Node Outages}
\label{sec:main_results}

\begin{table*}[t]
\centering
\tiny
\setlength{\tabcolsep}{2.25pt}
\renewcommand{\arraystretch}{1.08}
\caption{
Reconstruction under 30\% and 60\% long-horizon node outages. Entries report MAE / CRPS / Peak-$\Delta$MAE. MAE and Peak-$\Delta$MAE are in original units; CRPS is in standardized space. PeMS08 reports speed. ``--'' denotes deterministic methods. Lower is better; best results are in \textbf{bold}.
}

\label{tab:main_results}

\textbf{(a) Traffic datasets}\\[2pt]
\resizebox{\textwidth}{!}{%
\begin{tabular}{lcccc}
\toprule
& \multicolumn{2}{c}{\textbf{PeMS08 Speed}}
& \multicolumn{2}{c}{\textbf{METR-LA Speed}} \\
\cmidrule(lr){2-3}
\cmidrule(lr){4-5}
\textbf{Method}
& \textbf{30\%}
& \textbf{60\%}
& \textbf{30\%}
& \textbf{60\%} \\
\midrule

Graph Neighbor
& 3.498 / -- / 0.374
& 4.850 / -- / 0.524
& 5.956 / -- / 0.568
& 6.987 / -- / 0.668 \\

SAITS
& 3.148 / -- / 0.331
& 4.160 / -- / 0.418
& 4.982 / -- / 0.458
& 6.985 / -- / 0.598 \\

GRIN
& 2.889 / -- / 0.309
& 3.450 / -- / 0.365
& 3.423 / -- / 0.321
& 4.555 / -- / 0.398 \\

OPCR
& 2.558 / -- / 0.279
& 3.150 / -- / 0.318
& 2.612 / -- / 0.288
& 3.141 / -- / 0.323 \\

CSDI
& 3.089 / 0.0370 / 0.321
& 3.780 / 0.0428 / 0.375
& 3.851 / 0.0418 / 0.374
& 4.551 / 0.0498 / 0.471 \\

PriSTI
& 2.848 / 0.0340 / 0.295
& 3.440 / 0.0397 / 0.341
& 3.721 / 0.0363 / 0.342
& 3.853 / 0.0496 / 0.454 \\

CDSTE
& 2.499 / 0.0300 / 0.269
& 2.742 / 0.0334 / 0.294
& 2.981 / \textbf{0.0294} / 0.314
& 3.065 / \textbf{0.0345} / 0.317 \\

GiFlow
& 2.795 / 0.0330 / 0.305
& 3.210 / 0.0370 / 0.343
& 2.696 / 0.0334 / 0.341
& 2.754 / 0.0352 / 0.356 \\

\midrule

FM-TrainStruct
& 2.648 / 0.0310 / 0.299
& 3.325 / 0.0370 / 0.345
& 2.942 / 0.0295 / 0.316
& 3.742 / 0.0356 / 0.384 \\

Fixed-CTRL
& 2.389 / 0.0290 / 0.245
&  2.938 / 0.0324 / 0.295
& 2.933 / 0.0343 / 0.353
& 3.226 / 0.0381 / 0.384 \\
 
\textbf{CTRL-FM}
& \textbf{2.137} / \textbf{0.0260} / \textbf{0.209}
& \textbf{2.559} / \textbf{0.0310} / \textbf{0.249}
& \textbf{2.271} / 0.0314 / \textbf{0.256}
& \textbf{2.512} / 0.0355 / \textbf{0.294} \\

\bottomrule
\end{tabular}%
}

\vspace{5pt}

\textbf{(b) Air-quality datasets}\\[2pt]
\resizebox{\textwidth}{!}{%
\begin{tabular}{lcccc}
\toprule
& \multicolumn{2}{c}{\textbf{Air-36 PM$_{2.5}$}}
& \multicolumn{2}{c}{\textbf{AQI PM$_{2.5}$}} \\
\cmidrule(lr){2-3}
\cmidrule(lr){4-5}
\textbf{Method}
& \textbf{30\%}
& \textbf{60\%}
& \textbf{30\%}
& \textbf{60\%} \\
\midrule

Graph Neighbor
& 22.628 / -- / 3.856
& 29.938 / -- / 3.884
& 21.356 / -- / 3.144
& 28.589 / -- / 3.658 \\

SAITS
& 19.624 / -- / 2.761
& 25.634 / -- / 3.394
& 20.684 / -- / 3.024
& 28.695 / -- / 3.648 \\

GRIN
& 17.281 / -- / 2.424
& 23.629 / -- / 2.988
& 17.550 / -- / 2.569
& 20.278 / -- / 2.866 \\

OPCR
& \textbf{11.254} / -- / 2.161
& 18.265 / -- / 2.785
& 13.112 / -- / 2.018
& 15.226 / -- / 2.481 \\

CSDI
& 17.321 / 0.1360 / 2.860
& 24.831 / 0.1560 / 3.140
& 16.325 / 0.1200 / 2.336
& 22.354 / 0.1650 / 2.839 \\

PriSTI
& 17.520 / 0.1250 / 2.730
& 23.658 / 0.1380 / 2.960
& 16.050 / 0.0996 / 2.683
& 20.270 / 0.1500 / 2.758 \\

GiFlow
& 15.180 / 0.1090 / 2.256
& 19.704 / 0.1220 / 2.283
& \textbf{10.902} / 0.0918 / 2.003
& 16.022 / 0.1240 / 2.364 \\

\midrule

FM-TrainStruct
& 16.920 / 0.1040 / 2.466
& 22.280 / 0.1370 / 2.870
& 15.199 / 0.0921 / 2.580
& 20.486 / 0.1030 / 2.980 \\

Fixed-CTRL
& 14.580 / 0.0950 / 2.095
& 17.350 / 0.1118 / 2.244
& 12.990 / 0.0910 / 1.921
& 14.250 / 0.1130 / 2.310 \\

\textbf{CTRL-FM}
& 13.190 / \textbf{0.0790} / \textbf{1.980}
& \textbf{16.381} / \textbf{0.0910} / \textbf{2.187}
& 11.654 / \textbf{0.0896} / \textbf{1.865}
& \textbf{12.490} / \textbf{0.1020} / \textbf{2.153} \\

\bottomrule
\end{tabular}%
}
\end{table*}

\paragraph{Main results.}
Table~\ref{tab:main_results} reports the mean performance over three random
seeds. CTRL-FM delivers the strongest overall results across all datasets and
outage levels. On PeMS08, it achieves the lowest MAE, CRPS, and
Peak-$\Delta$MAE at both 30\% and 60\% outage. On METR-LA, it obtains the
best MAE and Peak-$\Delta$MAE under both settings, while CDSTE retains the
lowest CRPS. For Air-36 and AQI at 30\% outage, OPCR and GiFlow achieve the
lowest MAE, respectively, whereas CTRL-FM provides the best CRPS and
Peak-$\Delta$MAE. Under the more severe 60\% outage, CTRL-FM ranks first
across all reported metrics on both air-quality datasets. These results show
that outage-conditioned structural control is particularly effective as the
available sensor support decreases. Complete results, additional baselines,
and mean $\pm$ standard deviation values are reported in Appendix
Section~2.6; the main table reports means only.

% ============================================================
\subsection{Ablation and Inference Efficiency}
\label{sec:ablation_efficiency}

\begin{table*}[t]
\centering
\tiny
\setlength{\tabcolsep}{2.3pt}
\renewcommand{\arraystretch}{1.08}

\caption{
Component ablation under 60\% long-horizon node outage. Dataset entries
report MAE / Peak-$\Delta$MAE, with PeMS08 evaluated on speed. Runtime and
memory are normalized to FM-Base. Lower values are better, and the best
reconstruction value for each dataset and metric is shown in
\textbf{bold}.
}
\label{tab:component_ablation}

\resizebox{\textwidth}{!}{%
\begin{tabular}{lcccc|cc}
\toprule
\textbf{Variant}
& \textbf{PeMS08 Speed}
& \textbf{METR-LA Speed}
& \textbf{Air-36 PM$_{2.5}$}
& \textbf{AQI PM$_{2.5}$}
& \textbf{Time$\downarrow$}
& \textbf{Mem.$\downarrow$} \\
\midrule

FM-Base
& 3.853 / 0.393
& 3.960 / 0.417
& 28.548 / 3.709
& 25.927 / 3.813
& 1.00$\times$
& 1.00$\times$ \\

Fixed-CTRL
&  2.938 / 0.295
& 3.226 / 0.384
& 17.350 / 2.244
& 14.250 / 2.310
& 1.08$\times$
& 1.03$\times$ \\

CTRL-FM w/o reliability
& 2.796 / 0.252
& 2.818 / 0.299
& 16.758 / 2.195
& 13.210 / 2.223
& 1.15$\times$
& 1.04$\times$ \\

CTRL-FM w/o abstention
& 2.830 / 0.271
& 2.936 / 0.312
& 16.720 / 2.232
& 13.666 / 2.265
& 1.16$\times$
& 1.05$\times$ \\

\midrule

\textbf{CTRL-FM}
& \textbf{2.559} / \textbf{0.249}
& \textbf{2.512} / \textbf{0.294}
& \textbf{16.381} / \textbf{2.187}
& \textbf{12.490} / \textbf{2.153}
& 1.16$\times$
& 1.06$\times$ \\

\bottomrule
\end{tabular}%
}
\end{table*}

Table~\ref{tab:component_ablation} shows that each evaluated component contributes to
CTRL-FM. Fixed structural control substantially improves over FM-Base, while
removing reliability modulation or abstention consistently degrades MAE and
Peak-$\Delta$MAE across all datasets. The complete CTRL-FM achieves the best
reconstruction results in every setting, confirming the complementary value
of outage-specific reliability and conservative policy rejection. These
gains require only modest overhead, with $1.16\times$ runtime and
$1.06\times$ memory relative to FM-Base.

% ============================================================
\subsection{Extreme-Outage and Qualitative Analysis}
\label{sec:robustness_qualitative}

\paragraph{Extreme-outage robustness.}
CTRL-FM remains the strongest of the evaluated controlled flow-matching variants on PeMS08, METR-LA, and AQI under 65\% and 75\% node outages. At 75\% outage, valid-proxy coverage remains
between 84.8\% and 89.3\%, despite the limited surviving support. These
results demonstrate graceful degradation under severe sensor loss. Complete
stress-test results are provided in Appendix Section 2.7.

\paragraph{Qualitative reconstruction.}
Figure~\ref{fig:qualitative_node_outage} shows that CTRL-FM most consistently
follows the ground-truth trajectories across flow, occupancy, and speed. It
recovers both smooth demand patterns and congestion-related speed reductions,
while its predictive intervals remain concentrated during stable periods and
expand around abrupt transitions.

\paragraph{Limitations.}
Although CTRL-FM achieves the strongest overall performance among the
evaluated methods and degrades more gracefully under severe outages, it
still relies on sufficient surviving observations to construct informative
proxies and on a sensor graph that captures meaningful dependencies.
Performance may decline under extremely sparse or highly clustered failures.
Future work should consider real outage events, larger networks, and adaptive
graph structures.
\FloatBarrier
\section{Conclusion}
\label{sec:conclusion}

We introduced CTRL-FM, an observation-preserving flow-matching framework for long-horizon node outages that treats structural guidance as an outage-conditioned inference-time control problem rather than a fixed modeling assumption. Unlike approaches that impose the same graph and temporal constraints across different failure scenarios, CTRL-FM selects and adapts structural guidance based on the available observations and the characteristics of each realized outage. By separating reconstruction learning from inference-time structural control, the framework enables a frozen flow-matching backbone to adapt its sampling dynamics without retraining or modifying learned parameters. Specifically, self-supervised local proxy outages provide an evidence-based mechanism for evaluating candidate control policies, while observed-only reliability estimation and boundary-anchored guidance determine where graph and temporal information can be trusted. Conservative abstention further prevents structural intervention when proxy evidence does not sufficiently support improvement over uncontrolled sampling. Throughout this process, structural corrections remain restricted to missing entries, ensuring exact preservation of observed measurements.

Experiments on four traffic and air-quality datasets demonstrate the effectiveness of CTRL-FM under 30\% and 60\% long-horizon node outages, achieving the strongest overall reconstruction performance, particularly under severe sensor loss. Proxy-selection analysis demonstrates that surviving observations provide informative signals for identifying suitable control policies, while ablation studies confirm the complementary benefits of reliability modulation and conservative abstention. Additional stress tests with up to 75\% node outages show graceful performance degradation, with only modest computational overhead relative to the uncontrolled flow-matching backbone. Nevertheless, the framework relies on sufficiently informative surviving observations and meaningful spatial relationships, which may limit its effectiveness under extremely sparse or spatially clustered failures. Future work could explore adaptive proxy construction, dynamically evolving graph structures, and validation on real-world sensor outage events. More broadly, these findings suggest that structural guidance in generative reconstruction models can be validated and adapted at inference time rather than imposed uniformly across all scenarios, offering a promising direction toward more reliable and robust virtual sensing systems.
\section{Acknowledgments}
This work was supported by the Center for Smart Streetscapes, an NSF Engineering Research Center,
under grant agreement EEC-2133516

\bibliographystyle{unsrtnat}
\bibliography{references}  %%% Uncomment this line and comment out the ``thebibliography'' section below to use the external .bib file (using bibtex) .

@inproceedings{Khalili2026STREAMOOD,
  author    = {Khalili, B. and Bardey, A. and Smyth, A. W.},
  title     = {{STREAM-OOD}: Regime-Aware Sequential Monitoring for Streaming Out-of-Distribution Detection},
  booktitle = {Proceedings of the IEEE/CVF Conference on Computer Vision and Pattern Recognition},
  pages     = {7732--7741},
  year      = {2026}
}

@inproceedings{Yang2025GraphStructureLearningImputation,
  author    = {Xinyu Yang and Yu Sun and Xinyang Chen and Ying Zhang and Xiaojie Yuan},
  title     = {Graph Structure Learning for Spatial-Temporal Imputation:
               Adapting to Node and Feature Scales},
  booktitle = {Proceedings of the AAAI Conference on Artificial Intelligence},
  year      = {2025},
  volume    = {39},
  number    = {1},
  pages     = {959--967},
  doi       = {10.1609/aaai.v39i1.32081}
}

@article{Chen2025FourthOrderTensorCompletion,
  author  = {H. Chen and M. Lin and L. Zhao and Z. Xu and X. Luo},
  title   = {Fourth-Order Dimension Preserved Tensor Completion with
             Temporal Constraint for Missing Traffic Data Imputation},
  journal = {IEEE Transactions on Intelligent Transportation Systems},
  year    = {2025},
  volume  = {26},
  number  = {5},
  pages   = {6734--6748},
  doi     = {10.1109/TITS.2025.3531221}
}

@inproceedings{Ye2021GATTrafficImputation,
  author    = {Yongchao Ye and Shiyao Zhang and James J. Q. Yu},
  title     = {Spatial-Temporal Traffic Data Imputation via Graph
               Attention Convolutional Network},
  booktitle = {Artificial Neural Networks and Machine Learning -- ICANN 2021},
  year      = {2021},
  pages     = {241--252},
  address   = {Cham},
  publisher = {Springer},
  doi       = {10.1007/978-3-030-86362-3_20}
}

@article{Shen2023BidirectionalGATRecurrentImputation,
  author  = {Guojiang Shen and Wenfeng Zhou and Wenyi Zhang and
             Nali Liu and Zhi Liu and Xiangjie Kong},
  title   = {Bidirectional Spatial--Temporal Traffic Data Imputation
             via Graph Attention Recurrent Neural Network},
  journal = {Neurocomputing},
  year    = {2023},
  volume  = {531},
  pages   = {151--162},
  doi     = {10.1016/j.neucom.2023.02.017}
}

@article{Jia2025ReviewGNNProcessMonitoring,
  author  = {Mingwei Jia and Yuan Yao and Yi Liu},
  title   = {Review on Graph Neural Networks for Process Soft Sensor
             Development, Fault Diagnosis, and Process Monitoring},
  journal = {Industrial \& Engineering Chemistry Research},
  year    = {2025},
  volume  = {64},
  number  = {17},
  pages   = {8543--8564},
  doi     = {10.1021/acs.iecr.5c00283}
}

@article{He2025SpatiotemporalGCNWaterLevel,
  author  = {Li He and Jun Nan and Lei Chen and Xuesong Ye and
             Shasha Ji and Zewei Chen and Yibo Zhang and Fangmin Wu and
             Bohan Liu and Zhencheng Ge and Yanhan Che},
  title   = {Spatiotemporal Graph Convolutional Network Using Sparse
             Monitoring Data for Accurate Water-Level Reconstruction
             in Urban Drainage Systems},
  journal = {Journal of Hydrology},
  year    = {2025},
  volume  = {652},
  pages   = {132681},
  doi     = {10.1016/j.jhydrol.2025.132681}
}

@inproceedings{Zhong2023IterativeGraphReconstructionImputation,
  author    = {Jiajun Zhong and Ning Gui and Weiwei Ye},
  title     = {Data Imputation with Iterative Graph Reconstruction},
  booktitle = {Proceedings of the AAAI Conference on Artificial Intelligence},
  year      = {2023},
  volume    = {37},
  number    = {9},
  pages     = {11399--11407},
  doi       = {10.1609/aaai.v37i9.26348}
}

@article{Xu2024HierarchicalSTGCNImputation,
  author  = {Dong Xu and Hong Peng and Yu Tang and Hongbin Guo},
  title   = {Hierarchical Spatio-Temporal Graph Convolutional Neural
             Networks for Traffic Data Imputation},
  journal = {Information Fusion},
  year    = {2024},
  volume  = {106},
  pages   = {102292}
}

@inproceedings{Cuza2022StochasticTrafficSpeedImputation,
  author    = {Cuza, Carlos Enrique Muniz and Ho, Nguyen and
               Zacharatou, Eleni Tzirita and Pedersen, Torben Bach and
               Yang, Bin},
  title     = {Spatio-Temporal Graph Convolutional Network for
               Stochastic Traffic Speed Imputation},
  booktitle = {Proceedings of the 30th ACM International Conference on
               Advances in Geographic Information Systems},
  year      = {2022},
  articleno = {14},
  pages     = {1--12},
  publisher = {Association for Computing Machinery},
  doi       = {10.1145/3557915.3560948}
}

@article{Lei2022BayesianKernelizedMFImputation,
  author  = {Mengying Lei and Aur{\'e}lie Labb{\'e} and
             Yuankai Wu and Lijun Sun},
  title   = {Bayesian Kernelized Matrix Factorization for
             Spatiotemporal Traffic Data Imputation and Kriging},
  journal = {IEEE Transactions on Intelligent Transportation Systems},
  year    = {2022},
  volume  = {23},
  number  = {10},
  pages   = {18962--18974}
}

@misc{Wu2021SpatialAggregationTemporalConvKriging,
  author        = {Yuankai Wu and Dingyi Zhuang and Mengying Lei and
                   Aur{\'e}lie Labb{\'e} and Lijun Sun},
  title         = {Spatial Aggregation and Temporal Convolution
                   Networks for Real-Time Kriging},
  year          = {2021},
  eprint        = {2109.12144},
  archivePrefix = {arXiv},
  primaryClass  = {cs.LG}
}

@inproceedings{Zhou2015SpatioTemporalTensorCompletion,
  author    = {Hao Zhou and Daqing Zhang and Kunyuan Xie and
               Yan Chen},
  title     = {Spatio-Temporal Tensor Completion for Imputing
               Missing Internet Traffic Data},
  booktitle = {Proceedings of the 34th IEEE International Performance
               Computing and Communications Conference},
  year      = {2015},
  pages     = {1--7},
  publisher = {IEEE}
}

@inproceedings{cao2018brits,
  author    = {Wei Cao and Dong Wang and Jian Li and Hao Zhou and
               Lei Li and Yitan Li},
  title     = {{BRITS}: Bidirectional Recurrent Imputation for
               Time Series},
  booktitle = {Advances in Neural Information Processing Systems},
  year      = {2018},
  volume    = {31}
}

@article{du2023saits,
  author  = {Wenjie Du and David C{\^o}t{\'e} and Yan Liu},
  title   = {{SAITS}: Self-Attention-Based Imputation for Time Series},
  journal = {Expert Systems with Applications},
  year    = {2023},
  volume  = {219},
  pages   = {119619},
  doi     = {10.1016/j.eswa.2023.119619}
}

@inproceedings{wu2021ignnk,
  author    = {Yuankai Wu and Dingyi Zhuang and
               Aur{\'e}lie Labb{\'e} and Lijun Sun},
  title     = {Inductive Graph Neural Networks for
               Spatiotemporal Kriging},
  booktitle = {Proceedings of the AAAI Conference on Artificial Intelligence},
  year      = {2021},
  volume    = {35},
  number    = {5},
  pages     = {4478--4485},
  doi       = {10.1609/aaai.v35i5.16575}
}

@inproceedings{appleby2020kcn,
  author    = {Gabriel Appleby and Linfeng Liu and Li-Ping Liu},
  title     = {Kriging Convolutional Networks},
  booktitle = {Proceedings of the AAAI Conference on Artificial Intelligence},
  year      = {2020},
  volume    = {34},
  number    = {4},
  pages     = {3187--3194},
  doi       = {10.1609/aaai.v34i04.5716}
}

@inproceedings{cini2021grin,
  author    = {Andrea Cini and Ivan Marisca and Cesare Alippi},
  title     = {Filling the Gaps: Multivariate Time Series Imputation
               by Graph Neural Networks},
  booktitle = {The Tenth International Conference on Learning Representations},
  year      = {2022}
}

@inproceedings{marisca2022spin,
  author    = {Ivan Marisca and Andrea Cini and Cesare Alippi},
  title     = {Learning to Reconstruct Missing Data from
               Spatiotemporal Graphs with Sparse Observations},
  booktitle = {Advances in Neural Information Processing Systems},
  year      = {2022},
  volume    = {35},
  pages     = {32069--32082},
  doi       = {10.52202/068431-2324}
}

@inproceedings{liu2024opcr,
  author    = {Leyan Deng and Defu Lian and Chenwang Wu and Enhong Chen},
  title     = {Learning from Highly Sparse Spatio-Temporal Data},
  booktitle = {Advances in Neural Information Processing Systems},
  year      = {2024},
  volume    = {37},
  pages     = {94022--94046},
  doi       = {10.52202/079017-2982}
}

@inproceedings{tashiro2021csdi,
  author    = {Yusuke Tashiro and Jiaming Song and Yang Song and
               Stefano Ermon},
  title     = {{CSDI}: Conditional Score-Based Diffusion Models for
               Probabilistic Time Series Imputation},
  booktitle = {Advances in Neural Information Processing Systems},
  year      = {2021},
  volume    = {34},
  pages     = {24804--24816}
}

@inproceedings{liu2023pristi,
  author    = {Mingzhe Liu and Han Huang and Hao Feng and Leilei Sun and
               Bowen Du and Yanjie Fu},
  title     = {{PriSTI}: A Conditional Diffusion Framework for
               Spatiotemporal Imputation},
  booktitle = {Proceedings of the 39th IEEE International Conference
               on Data Engineering},
  year      = {2023},
  pages     = {1927--1939},
  publisher = {IEEE},
  doi       = {10.1109/ICDE55515.2023.00150}
}

@article{lei2024cdste,
  author  = {Da Lei and Min Xu and Shuaian Wang},
  title   = {A Conditional Diffusion Model for Probabilistic
             Estimation of Traffic States at Sensor-Free Locations},
  journal = {Transportation Research Part C: Emerging Technologies},
  year    = {2024},
  volume  = {166},
  pages   = {104798},
  doi     = {10.1016/j.trc.2024.104798}
}

@inproceedings{Zhang2025Diffputer,
  author    = {Hao Zhang and Liyue Fang and Qitian Wu and Philip S. Yu},
  title     = {Diffputer: Empowering Diffusion Models for
               Missing Data Imputation},
  booktitle = {The Thirteenth International Conference on
               Learning Representations},
  year      = {2025}
}

@inproceedings{Gao2025SSDTS,
  author    = {Hongfan Gao and Weijie Shen and Xiaofeng Qiu and
               Rui Xu and Bin Yang and Jia Hu},
  title     = {{SSD-TS}: Exploring the Potential of Linear State
               Space Models for Diffusion Models in Time Series Imputation},
  booktitle = {Proceedings of the 31st ACM SIGKDD Conference on
               Knowledge Discovery and Data Mining},
  year      = {2025},
  volume    = {2},
  pages     = {649--660}
}

@inproceedings{Lipman2022FlowMatching,
  author    = {Yaron Lipman and Ricky T. Q. Chen and Heli Ben-Hamu and
               Maximilian Nickel and Matt Le},
  title     = {Flow Matching for Generative Modeling},
  booktitle = {The Eleventh International Conference on
               Learning Representations},
  year      = {2023},
  eprint    = {2210.02747},
  archivePrefix = {arXiv},
  primaryClass  = {cs.LG}
}

@inproceedings{Wang2024TrainingFreeGuidedFlowMatching,
  author    = {Luran Wang and Chaoran Cheng and Yizhen Liao and
               Yanru Qu and Ge Liu},
  title     = {Training Free Guided Flow-Matching with Optimal Control},
  booktitle = {The Thirteenth International Conference on
               Learning Representations},
  year      = {2025},
  eprint    = {2410.18070},
  archivePrefix = {arXiv},
  primaryClass  = {cs.LG}
}

@inproceedings{WanElucidatingFlowMatchingDynamics,
  author    = {Zhengchao Wan and Qingsong Wang and
               Gal Mishne and Yusu Wang},
  title     = {Elucidating Flow Matching {ODE} Dynamics via
               Data Geometry and Denoisers},
  booktitle = {Proceedings of the 42nd International Conference
               on Machine Learning},
  year      = {2025},
  volume    = {267},
  series    = {Proceedings of Machine Learning Research},
  pages     = {62020--62083},
  publisher = {PMLR},
  url       = {https://proceedings.mlr.press/v267/wan25e.html}
}

@misc{Xu2025EnergyGuidedFlowMatching,
  author        = {Guikun Xu and Xiaohan Yi and Peilin Zhao and Yatao Bian},
  title         = {Energy-Guided Flow Matching Enables Few-Step
                   Conformer Generation and Ground-State Identification},
  year          = {2025},
  eprint        = {2512.22597},
  archivePrefix = {arXiv},
  primaryClass  = {cs.LG}
}

@inproceedings{Zhang2025HierarchicalRectifiedFlow,
  author    = {Yichi Zhang and Yici Yan and Alex Schwing and Zhizhen Zhao},
  title     = {Towards Hierarchical Rectified Flow},
  booktitle = {The Thirteenth International Conference on
               Learning Representations},
  year      = {2025},
  eprint    = {2502.17436},
  archivePrefix = {arXiv},
  primaryClass  = {cs.LG}
}

@inproceedings{giflow2025,
  author    = {Zepeng Zhang and Aref Einizade and
               Jhony H. Giraldo and Olga Fink},
  title     = {Spatiotemporal Imputation with
               Graph-Informed Flow Matching},
  booktitle = {Proceedings of the 43rd International Conference
               on Machine Learning},
  year      = {2026},
  note      = {Accepted at ICML 2026},
  eprint    = {2606.06682},
  archivePrefix = {arXiv},
  primaryClass  = {cs.LG}
}

@article{gneiting2007strictly,
  author  = {Tilmann Gneiting and Adrian E. Raftery},
  title   = {Strictly Proper Scoring Rules, Prediction, and Estimation},
  journal = {Journal of the American Statistical Association},
  year    = {2007},
  volume  = {102},
  number  = {477},
  pages   = {359--378},
  doi     = {10.1198/016214506000001437}
}

@article{Chen2001PeMS,
  author  = {Chen, Chao and Petty, Karl and Skabardonis, Alexander
             and Varaiya, Pravin and Jia, Zhanfeng},
  title   = {Freeway Performance Measurement System:
             Mining Loop Detector Data},
  journal = {Transportation Research Record},
  volume  = {1748},
  number  = {1},
  pages   = {96--102},
  year    = {2001},
  doi     = {10.3141/1748-12}
}

@inproceedings{Guo2019ASTGCN,
  author    = {Guo, Shengnan and Lin, Youfang and Feng, Ning
               and Song, Chao and Wan, Huaiyu},
  title     = {Attention Based Spatial-Temporal Graph Convolutional
               Networks for Traffic Flow Forecasting},
  booktitle = {Proceedings of the AAAI Conference on Artificial Intelligence},
  volume    = {33},
  number    = {01},
  pages     = {922--929},
  year      = {2019},
  doi       = {10.1609/aaai.v33i01.3301922}
}

@inproceedings{Li2018DCRNN,
  author    = {Li, Yaguang and Yu, Rose and Shahabi, Cyrus and Liu, Yan},
  title     = {Diffusion Convolutional Recurrent Neural Network:
               Data-Driven Traffic Forecasting},
  booktitle = {International Conference on Learning Representations},
  year      = {2018}
}

@inproceedings{Yi2016STMVL,
  author    = {Yi, Xiuwen and Zheng, Yu and Zhang, Junbo and Li, Tianrui},
  title     = {{ST-MVL}: Filling Missing Values in Geo-Sensory
               Time Series Data},
  booktitle = {Proceedings of the Twenty-Fifth International Joint
               Conference on Artificial Intelligence},
  pages     = {2704--2710},
  year      = {2016}
}

@inproceedings{Zheng2015AirQuality,
  author    = {Zheng, Yu and Yi, Xiuwen and Li, Ming and Li, Rui
               and Shan, Zhangqing and Chang, Eric and Li, Tianrui},
  title     = {Forecasting Fine-Grained Air Quality Based on Big Data},
  booktitle = {Proceedings of the 21st ACM SIGKDD International
               Conference on Knowledge Discovery and Data Mining},
  pages     = {2267--2276},
  year      = {2015},
  doi       = {10.1145/2783258.2788573}
}

% ============================================================

\clearpage
\section*{Appendix}

% Restart section numbering from 1
\setcounter{section}{0}
\setcounter{subsection}{0}
\setcounter{subsubsection}{0}

\section{Additional Method Details}
\label{app:method_details}

This supplementary material provides the complete policy library,
proxy-calibration procedure, boundary-anchored guidance construction,
observed-only reliability estimation, observation-preservation proof, and
additional implementation details for CTRL-FM. All backbone and reliability
parameters are trained on the training split and frozen before test-time
calibration.

% ============================================================
\subsection{Policy Library and Proxy Calibration}
\label{app:policy_proxy_details}

CTRL-FM selects how the frozen flow field should be controlled separately for
each realized outage. The candidate policies are fixed before test
evaluation, while their local evaluation is performed using self-supervised
proxy outages constructed from surviving measurements.

\subsection{Candidate Policy Library}

The policy library is
\begin{equation}
\mathcal H
=
\left\{
h_r=
(\alpha_r,\lambda_{g,r},\lambda_{t,r})
\right\}_{r=0}^{R},
\label{eq:app_controller_library}
\end{equation}
where \(\alpha_r\) controls the total structural correction and
\(\lambda_{g,r}\) and \(\lambda_{t,r}\) determine its graph--temporal
balance.

The uncontrolled policy is
\begin{equation}
h_0=(0,0,0),
\label{eq:app_uncontrolled_policy}
\end{equation}
while every controlled policy satisfies
\begin{equation}
\alpha_r>0,
\qquad
\lambda_{g,r},\lambda_{t,r}\geq0,
\qquad
\lambda_{g,r}+\lambda_{t,r}=1.
\label{eq:app_controlled_policy_constraints}
\end{equation}

The candidate library used in our experiments is summarized in
Table~\ref{tab:app_policy_library}.

The library is globally shared and selected using validation outages.
However, policy evaluation and selection are repeated locally for every
realized test outage. The policies therefore represent alternative ways of
controlling the same frozen flow-matching backbone rather than separately
trained reconstruction models.
\FloatBarrier
\subsection{Outage-Conditioned Proxy Construction}

Let \(\mathcal O\subseteq V\) denote the true failed-node set and
\(\mathcal I\) its outage interval. The true outage remains active throughout
calibration, and its unavailable values are never accessed.

CTRL-FM constructs \(B\) proxy outages by temporarily hiding nearby
surviving measurements. Each proxy provides a local self-supervised
evaluation set for comparing the candidate policies.

\paragraph{Eligible proxy nodes.}
The distance from a surviving node \(j\) to the failed set is
\begin{equation}
d_G(j,\mathcal O)
=
\min_{k\in\mathcal O}d_G(j,k),
\label{eq:app_graph_distance_to_outage}
\end{equation}
where \(d_G(j,k)\) denotes shortest-path graph distance.

For a candidate interval \(\mathcal J\), the eligible proxy nodes are
\begin{equation}
\mathcal V_{\mathrm{proxy}}(\mathcal J)
=
\left\{
j\in V\setminus\mathcal O:
\begin{array}{l}
M_{t,j,:}=\mathbf 1
\ \text{for all }t\in\mathcal J,\\
d_G(j,\mathcal O)\leq r_p
\end{array}
\right\}.
\label{eq:app_proxy_candidate_nodes}
\end{equation}
Thus, each eligible node is close to the true outage and fully observed over
the proxy interval.

\paragraph{Proxy nodes and intervals.}
Proxy \(b\) contains a node set \(\mathcal O^{(b)}\) and an interval
\(\mathcal I^{(b)}\) satisfying
\begin{equation}
\mathcal O^{(b)}
\subseteq
\mathcal V_{\mathrm{proxy}}
\left(
\mathcal I^{(b)}
\right),
\qquad
\mathcal I^{(b)}\subseteq\mathcal I.
\label{eq:app_proxy_components}
\end{equation}

The proxy node budget is
\begin{equation}
\left|\mathcal O^{(b)}\right|
=
\min
\left(
m_p,\,
\left|
\mathcal V_{\mathrm{proxy}}
\left(
\mathcal I^{(b)}
\right)
\right|
\right),
\label{eq:app_proxy_size}
\end{equation}
The proxy-node budget is \(m_p=4\), so each proxy temporarily
hides at most four eligible surviving nodes.

The proxy interval length is
\begin{equation}
\left|\mathcal I^{(b)}\right|
=
\min
\left(
|\mathcal I|,
L_{\mathrm{probe}}
\right).
\label{eq:app_proxy_interval_length}
\end{equation}
For outages longer than \(L_{\mathrm{probe}}\), different proxies may cover
different subintervals of \(\mathcal I\). Node overlap is minimized when
sufficient eligible nodes are available.

\paragraph{Proxy masks.}
The same node-level proxy mask is applied across all channels:
\begin{equation}
P^{(b)}_{t,k,c}
=
\mathbb I
\left[
t\in\mathcal I^{(b)}
\ \text{and}\
k\in\mathcal O^{(b)}
\right].
\label{eq:app_proxy_mask}
\end{equation}

A valid proxy satisfies
\begin{equation}
\left\|P^{(b)}\right\|_1>0,
\qquad
P^{(b)}\odot U=0.
\label{eq:app_proxy_observed_constraint}
\end{equation}
The second condition guarantees that every proxy target was originally
observed and does not overlap the true outage.

The temporary observation and missing masks are
\begin{equation}
M^{(b)}
=
M\odot\left(1-P^{(b)}\right),
\qquad
U^{(b)}
=
1-M^{(b)}
=
U+P^{(b)}.
\label{eq:app_temporary_proxy_masks}
\end{equation}
\begin{table}[t]
\centering
\small
\setlength{\tabcolsep}{4pt}
\renewcommand{\arraystretch}{1.05}
\caption{
Candidate structural-control policies. The parameter \(\alpha\) controls
the total correction strength, while \(\lambda_g\) and \(\lambda_t\)
determine the graph--temporal balance.
}
\label{tab:app_policy_library}
\begin{tabular}{lccc}
\toprule
\textbf{Policy}
& \(\boldsymbol{\alpha}\)
& \(\boldsymbol{\lambda_g}\)
& \(\boldsymbol{\lambda_t}\) \\
\midrule
\(h_0\): uncontrolled
& 0.00 & 0.00 & 0.00 \\
\(h_1\): weak balanced
& 0.50 & 0.50 & 0.50 \\
\(h_2\): graph-dominant
& 0.75 & 0.75 & 0.25 \\
\(h_3\): temporal-dominant
& 0.75 & 0.25 & 0.75 \\
\(h_4\): strong balanced
& 1.00 & 0.50 & 0.50 \\
\bottomrule
\end{tabular}
\end{table}
All boundary trends, structural guidance fields, observed-context features,
and reliability maps are recomputed using \(M^{(b)}\). A temporarily hidden
proxy value therefore cannot contribute to its own reconstruction or
reliability estimate.

\paragraph{Distance-based weighting.}
Let \(\mathcal B\) denote the set of retained valid proxies. The mean graph
distance of proxy \(b\) from the true outage is
\begin{equation}
d_b
=
\frac{1}{|\mathcal O^{(b)}|}
\sum_{j\in\mathcal O^{(b)}}
d_G(j,\mathcal O).
\label{eq:app_proxy_distance}
\end{equation}

Its normalized weight is
\begin{equation}
\omega_b
=
\frac{
\exp(-d_b/\tau_p)
}{
\displaystyle
\sum_{b'\in\mathcal B}
\exp(-d_{b'}/\tau_p)
},
\qquad
b\in\mathcal B.
\label{eq:app_proxy_weight}
\end{equation}
Nearby proxies receive greater weight because they better represent the
local observational context of the true outage.

\subsubsection{Proxy Calibration Objective}

All policies evaluated on the same proxy use identical Gaussian
initialization and the same reduced solver grid. This paired evaluation
reduces sampling variation and makes the policy scores directly comparable.

\paragraph{Proxy-value loss.}
For proxy \(b\) and policy \(h_r\), let
\(\widehat X^{(b,r)}\) denote the reconstruction produced by the reduced
proxy sampler. Its value error is
\begin{equation}
\mathcal L^{\mathrm{val}}_{b,r}
=
\frac{
\left\|
P^{(b)}
\odot
\left(
\widehat X^{(b,r)}-X
\right)
\right\|_1
}{
\|P^{(b)}\|_1+\epsilon_p
}.
\label{eq:app_proxy_value_loss}
\end{equation}
Only the temporarily hidden proxy entries are evaluated.

\paragraph{Temporal-change loss.}
For a signal \(Y\), define
\begin{equation}
(\Delta_tY)_{t,k,c}
=
Y_{t,k,c}-Y_{t-1,k,c},
\qquad
t=2,\ldots,T.
\label{eq:app_temporal_difference}
\end{equation}

A temporal difference is scored only when it touches a proxy entry and both
of its endpoints were originally observed:
\begin{equation}
Q^{(b)}_{t,k,c}
=
M_{t,k,c}M_{t-1,k,c}
\,
\mathbb I
\left[
P^{(b)}_{t,k,c}
+
P^{(b)}_{t-1,k,c}
>0
\right].
\label{eq:app_dynamic_proxy_mask}
\end{equation}

The dynamic error is
\begin{equation}
\mathcal L^{\mathrm{dyn}}_{b,r}
=
\frac{
\left\|
Q^{(b)}
\odot
\left(
\Delta_t\widehat X^{(b,r)}
-
\Delta_tX
\right)
\right\|_1
}{
\|Q^{(b)}\|_1+\epsilon_p
}.
\label{eq:app_proxy_dynamic_loss}
\end{equation}

The original-mask factors exclude every temporal difference that would
require access to a value inside the true outage. A proxy is retained only
when
\begin{equation}
\|P^{(b)}\|_1>0,
\qquad
\|Q^{(b)}\|_1>0.
\label{eq:app_valid_proxy_conditions}
\end{equation}

\paragraph{Complete proxy score.}
The score of policy \(h_r\) on proxy \(b\) is
\begin{equation}
\ell_{b,r}
=
\mathcal L^{\mathrm{val}}_{b,r}
+
\lambda_{\mathrm{dyn}}
\mathcal L^{\mathrm{dyn}}_{b,r}.
\label{eq:app_proxy_loss}
\end{equation}
The value term measures reconstruction accuracy, while the dynamic term
discourages policies that obtain low pointwise error by oversmoothing
temporal transitions.

The outage-level policy score is
\begin{equation}
\mathcal J_{\mathrm{proxy}}(h_r)
=
\sum_{b\in\mathcal B}
\omega_b\ell_{b,r}.
\label{eq:app_aggregate_proxy_loss}
\end{equation}
The locally preferred policy is
\begin{equation}
\widetilde h
=
\arg\min_{h_r\in\mathcal H}
\mathcal J_{\mathrm{proxy}}(h_r).
\label{eq:app_best_policy}
\end{equation}
Ties are resolved in favor of smaller guidance strength, with \(h_0\)
preferred when it ties with a controlled policy.

\paragraph{Conservative abstention.}
CTRL-FM measures the relative improvement over uncontrolled sampling:
\begin{equation}
\mathcal R(\widetilde h)
=
\frac{
\mathcal J_{\mathrm{proxy}}(h_0)
-
\mathcal J_{\mathrm{proxy}}(\widetilde h)
}{
\mathcal J_{\mathrm{proxy}}(h_0)
+
\epsilon_{\mathrm{abst}}
}.
\label{eq:app_relative_proxy_improvement}
\end{equation}

The final outage-specific policy is
\begin{equation}
h^\star
=
\begin{cases}
\widetilde h,
&
\mathcal R(\widetilde h)
\geq
\delta_{\mathrm{abst}},
\\[1mm]
h_0,
&
\text{otherwise}.
\end{cases}
\label{eq:app_abstention_rule}
\end{equation}

If no valid proxy exists or the improvement is insufficient, CTRL-FM
defaults to \(h_0\). No value inside the true outage is used for proxy
construction, reliability estimation, scoring, or policy selection.

% ============================================================
\subsection{Boundary-Anchored Structural Guidance}
\label{app:guidance_details}

CTRL-FM constructs graph and temporal guidance fields by comparing the
current missing-region flow velocity with valid trends extracted exclusively
from surviving measurements. These observed trends provide structural anchors
in the surviving region. Through graph compatibility, anchors adjacent to the
observed--missing boundary directly influence unavailable nodes, while all
observed measurements remain fixed.

All quantities in this section are expressed in the channel-wise standardized
model space. For multichannel PeMS08, flow, occupancy, and speed are
standardized independently using statistics computed from the training split.

Let \(M^\circ\) denote the active observation mask. It equals \(M\) during
final inference and \(M^{(b)}\) during proxy calibration. Define
\begin{equation}
U^\circ = 1-M^\circ.
\label{eq:app_active_missing_mask}
\end{equation}
The corresponding active state at flow time \(s\) is
\begin{equation}
X_{\mathrm{mix}}^\circ(s)
=
M^\circ\odot X
+
U^\circ\odot X_{\mathrm{gen}}(s).
\label{eq:app_active_mixed_state}
\end{equation}

\subsubsection{Boundary-Anchored Comparison Field}

Define the backward and forward validity indicators
\begin{equation}
\begin{aligned}
a^-_{t,k,c}
&=
M^\circ_{t,k,c}
M^\circ_{t-1,k,c},
\\
a^+_{t,k,c}
&=
M^\circ_{t,k,c}
M^\circ_{t+1,k,c},
\end{aligned}
\label{eq:app_valid_trend_pairs}
\end{equation}
with terms involving invalid temporal indices set to zero. Thus, a temporal
change is considered valid only when both measurements used to compute it are
available under the active mask.

The corresponding support mask is
\begin{equation}
Q_{t,k,c}
=
\mathbb I
\left[
a^-_{t,k,c}+a^+_{t,k,c}>0
\right].
\label{eq:app_valid_observed_trend}
\end{equation}

The observed physical-time trend is
\begin{equation}
d_{\mathrm{obs},t,k,c}
=
Q_{t,k,c}
\frac{
a^-_{t,k,c}
\left(
X_{t,k,c}-X_{t-1,k,c}
\right)
+
a^+_{t,k,c}
\left(
X_{t+1,k,c}-X_{t,k,c}
\right)
}{
\Delta
\max
\left\{
a^-_{t,k,c}+a^+_{t,k,c},
1
\right\}
},
\label{eq:app_observed_trend}
\end{equation}
where \(\Delta\) is the physical sampling interval. When only one neighboring
measurement is available, Eq.~\eqref{eq:app_observed_trend} uses the
corresponding one-sided difference. When both neighboring measurements are
available, it reduces to the centered difference
\[
\frac{X_{t+1,k,c}-X_{t-1,k,c}}{2\Delta}.
\]
Because all validity terms are derived from \(M^\circ\), no unavailable value
contributes to \(d_{\mathrm{obs}}\).

At flow time \(s\), define the detached structural comparison field
\begin{equation}
\overline A(s)
=
\operatorname{sg}
\left[
U^\circ\odot
v_\theta
\left(
X_{\mathrm{mix}}^\circ(s),
s,
G,
M^\circ
\right)
\right]
+
M^\circ\odot Q\odot
\kappa d_{\mathrm{obs}},
\label{eq:app_boundary_field}
\end{equation}
where \(\operatorname{sg}\) denotes stop-gradient and \(\kappa\) maps the
observed physical-time trend to the scale of the flow velocity.

The first term contains the detached backbone velocity on unavailable
entries. The second inserts measurement-supported trends on observed entries.
Together, they form a common comparison field in which the graph and temporal
compatibility of the current missing-region velocity can be evaluated against
valid surviving measurements. The observed components are used only for this
structural comparison and never update an observed measurement.

\subsection{Graph and Temporal Guidance Fields}
\label{app:guidance_fields}

For channel vectors \(a,b\in\mathbb R^C\), define the differentiable
channel-averaged discrepancy
\begin{equation}
d_\rho(a,b)
=
\frac{1}{C}
\sum_{c=1}^{C}
\left[
\sqrt{(a_c-b_c)^2+\epsilon_\rho^2}
-
\epsilon_\rho
\right],
\label{eq:app_smooth_difference}
\end{equation}
where \(\epsilon_\rho>0\) controls the smooth approximation to the absolute
difference. This discrepancy is nonnegative, equals zero when \(a=b\), and
has bounded derivatives.

For graph edge \((i,j)\), define its affinity as
\begin{equation}
w_{ij}
=
\exp
\left(
-\frac{\widetilde c_{ij}}{\tau_g}
\right),
\label{eq:app_graph_affinity}
\end{equation}
where \(\widetilde c_{ij}\geq 0\) is the normalized graph cost and
\(\tau_g>0\) is the affinity temperature. Consequently,
\(w_{ij}\in(0,1]\). For an undirected graph, each edge is included once in
the following summation.

The graph compatibility objective is
\begin{equation}
\mathcal C_g(A)
=
\frac{
\displaystyle
\sum_{t=1}^{T}
\sum_{(i,j)\in E}
w_{ij}
d_\rho
\left(
A_{t,i,:},
A_{t,j,:}
\right)
}{
\displaystyle
T
\sum_{(i,j)\in E}
w_{ij}
+
\epsilon_g
}.
\label{eq:app_graph_energy}
\end{equation}

The temporal compatibility objective is
\begin{equation}
\mathcal C_t(A)
=
\frac{
\displaystyle
\sum_{t=1}^{T-1}
\sum_{k=1}^{K}
d_\rho
\left(
A_{t+1,k,:},
A_{t,k,:}
\right)
}{
(T-1)K+\epsilon_t
}.
\label{eq:app_temporal_energy}
\end{equation}
This objective measures the weighted disagreement between velocity vectors at
connected sensors.
Unlike direct signal smoothing, this objective regularizes adjacent flow
velocities and therefore encourages dynamically coherent transport during
sampling.

The corresponding structural guidance fields are
\begin{equation}
g_G(s)
=
-
\left.
\nabla_A
\mathcal C_g(A)
\right|_{A=\overline A(s)},
\qquad
g_T(s)
=
-
\left.
\nabla_A
\mathcal C_t(A)
\right|_{A=\overline A(s)}.
\label{eq:app_guidance_fields}
\end{equation}

The negative gradients point toward lower graph and temporal compatibility
energies. The detached field \(\overline A(s)\) is treated as the
differentiation variable when evaluating these gradients. Because the
backbone velocity in Eq.~\eqref{eq:app_boundary_field} is wrapped by
stop-gradient, the backbone output is treated as fixed and no
vector--Jacobian product through \(v_\theta\) is required.

Although \(g_G(s)\) and \(g_T(s)\) are defined over the complete comparison
field, only their unavailable-region components affect reconstruction. The
controlled sampling velocity is projected by \(U^\circ\) during proxy
calibration and by \(U\) during final inference. Therefore, neither guidance
field can modify an observed measurement.

The fields in Eq.~\eqref{eq:app_guidance_fields} determine the graph and
temporal correction directions. Their outage-level strengths are specified by
the selected policy, while their local contributions are modulated by the
observed-only reliability maps \(R_G\) and \(R_T\), defined next.
% ============================================================
\subsection{Observed-Only Reliability Estimation}
\label{app:reliability_details}

Reliability is estimated exclusively from surviving measurements. The
features capture observational support, neighborhood agreement, change
magnitude, graph distance, and distance to an observed temporal boundary.

Let \(M^\circ\) denote the active mask, corresponding to the true test mask
\(M\), a temporary proxy mask \(M^{(b)}\), or a synthetic training mask
\(\widetilde M\). Let \(\bar X\) denote the channel-wise normalized signal.

\subsubsection{Observed Node Context}

For physical-time interval \((t,t+1)\), define
\begin{equation}
R_{t,j,c}
=
M^\circ_{t,j,c}
M^\circ_{t+1,j,c},
\qquad
t=1,\ldots,T-1,
\label{eq:app_valid_observed_change}
\end{equation}
and the corresponding valid observed change
\begin{equation}
D_{t,j,c}
=
R_{t,j,c}
\left(
\bar X_{t+1,j,c}
-
\bar X_{t,j,c}
\right).
\label{eq:app_observed_change}
\end{equation}
A temporal difference is therefore valid only when both endpoints are
observed.

Let
\begin{equation}
\mathcal N^+(k)
=
\{k\}\cup\mathcal N(k)
\label{eq:app_context_nodes}
\end{equation}
contain node \(k\) and its graph neighbors, and define
\begin{equation}
\overline w_{kj}
=
\begin{cases}
1, & j=k,\\
w_{kj}, & j\in\mathcal N(k).
\end{cases}
\label{eq:app_context_weight}
\end{equation}

\paragraph{Observed support.}
The available local observational support is
\begin{equation}
q_{k,t}
=
\frac{
\displaystyle
\sum_{j\in\mathcal N^+(k)}
\overline w_{kj}
\sum_{c=1}^{C}
R_{t,j,c}
}{
\displaystyle
C
\sum_{j\in\mathcal N^+(k)}
\overline w_{kj}
+
\epsilon_m
}.
\label{eq:app_context_support}
\end{equation}

\paragraph{Average observed change.}
For each channel, define
\begin{equation}
\mu_{k,t,c}
=
\frac{
\displaystyle
\sum_{j\in\mathcal N^+(k)}
\overline w_{kj}D_{t,j,c}
}{
\displaystyle
\sum_{j\in\mathcal N^+(k)}
\overline w_{kj}R_{t,j,c}
+
\epsilon_m
}.
\label{eq:app_context_mean_change}
\end{equation}

Its channel-averaged magnitude is
\begin{equation}
m_{k,t}
=
\frac{1}{C}
\left\|
\mu_{k,t,:}
\right\|_1.
\label{eq:app_context_change_magnitude}
\end{equation}

\paragraph{Observed disagreement.}
Local inconsistency among surviving sensors is measured by
\begin{equation}
v_{k,t}
=
\frac{
\displaystyle
\sum_{j\in\mathcal N^+(k)}
\overline w_{kj}
\sum_{c=1}^{C}
R_{t,j,c}
\left|
D_{t,j,c}-\mu_{k,t,c}
\right|
}{
\displaystyle
\sum_{j\in\mathcal N^+(k)}
\overline w_{kj}
\sum_{c=1}^{C}
R_{t,j,c}
+
\epsilon_m
}.
\label{eq:app_context_disagreement}
\end{equation}
Small values indicate coherent local dynamics, whereas large values indicate
disagreement among surviving sensors.

\paragraph{Distance to an observed boundary.}
Temporal support weakens deeper inside a long outage. We define
\begin{equation}
b_{k,t}
=
\min
\left(
1,\,
\frac{
\displaystyle
\min_{\tau:\,
\|M^\circ_{\tau,k,:}\|_1>0}
|t-\tau|
}{
L_{\max}
}
\right).
\label{eq:app_boundary_distance}
\end{equation}
If node \(k\) has no observation in the current window, we set
\(b_{k,t}=1\). We use \(L_{\max}=12\) steps, matching the selected proxy
calibration horizon. Thus, the boundary-distance feature saturates when the
nearest observed temporal boundary is at least 12 steps away.

All context quantities depend only on the active mask and surviving
measurements.

\subsubsection{Compact Reliability Features}

For graph edge \((i,j)\), define
\begin{equation}
\begin{aligned}
s^g_{ij,t}
&=
\frac{q_{i,t}+q_{j,t}}{2},
\\
d^g_{ij,t}
&=
\frac{1}{C}
\left\|
\mu_{i,t,:}-\mu_{j,t,:}
\right\|_1,
\\
v^g_{ij,t}
&=
\frac{v_{i,t}+v_{j,t}}{2},
\\
m^g_{ij,t}
&=
\frac{m_{i,t}+m_{j,t}}{2}.
\end{aligned}
\label{eq:app_graph_context_features}
\end{equation}

The graph reliability feature vector is
\begin{equation}
\phi_{ij,t}
=
\left[
s^g_{ij,t},
d^g_{ij,t},
v^g_{ij,t},
m^g_{ij,t},
\widetilde c_{ij}
\right].
\label{eq:app_graph_feature_vector}
\end{equation}
These features summarize observational support, agreement between endpoint
neighborhoods, within-neighborhood disagreement, local change magnitude, and
graph distance without requiring an unavailable endpoint value.

The temporal reliability feature vector is
\begin{equation}
\psi_{k,t}
=
\left[
q_{k,t},
m_{k,t},
v_{k,t},
b_{k,t}
\right].
\label{eq:app_temporal_feature_vector}
\end{equation}
All features are standardized using training-split statistics.

\subsubsection{Reliability Weights and Maps}

Let \(f_g\) and \(f_t\) be compact graph and temporal reliability networks:
\begin{equation}
\widehat\gamma_{ij,t}
=
\sigma
\left(
f_g(\phi_{ij,t})
\right),
\qquad
\widehat\eta_{k,t}
=
\sigma
\left(
f_t(\psi_{k,t})
\right).
\label{eq:app_reliability_predictions}
\end{equation}

The graph module \(f_g\) receives the five-dimensional feature vector
\(\phi_{ij,t}\). The temporal module \(f_t\)
receives the four-dimensional feature vector \(\psi_{k,t}\). Each module contains one hidden
layer with ReLU activation and dropout \(0.1\), followed by a scalar linear
output. The sigmoid in Eq.~\eqref{eq:app_reliability_predictions} maps the
output to \((0,1)\).

Their outputs are gated by observed support:
\begin{equation}
\gamma_{ij,t}
=
s^g_{ij,t}\widehat\gamma_{ij,t},
\qquad
\eta_{k,t}
=
q_{k,t}\widehat\eta_{k,t}.
\label{eq:app_reliability_weights}
\end{equation}
Reliability therefore becomes zero when no supporting observation is
available.

The reliability predictions are defined over physical-time intervals. To
align them with the time-indexed guidance fields, define the temporal
alignment operator
\begin{equation}
\left(\mathcal A_T z\right)_t
=
\begin{cases}
z_1,
&
t=1,
\\[1mm]
\dfrac{z_{t-1}+z_t}{2},
&
2\leq t\leq T-1,
\\[3mm]
z_{T-1},
&
t=T.
\end{cases}
\label{eq:app_temporal_alignment}
\end{equation}

The node-time graph reliability map is
\begin{equation}
(R_G)_{t,k,c}
=
\frac{
\displaystyle
\sum_{j\in\mathcal N(k)}
w_{kj}
\left(
\mathcal A_T\gamma_{kj}
\right)_t
}{
\displaystyle
\sum_{j\in\mathcal N(k)}
w_{kj}
+
\epsilon_g
}.
\label{eq:app_graph_reliability_map}
\end{equation}

The temporal reliability map is
\begin{equation}
(R_T)_{t,k,c}
=
\left(
\mathcal A_T\eta_k
\right)_t.
\label{eq:app_temporal_reliability_map}
\end{equation}

Both \(R_G\) and \(R_T\) have the same shape as their corresponding
guidance fields and can therefore directly modulate \(g_G\) and
\(g_T\) in the outage-conditioned controlled velocity defined in
Eq.~(20) of the main paper.

During proxy evaluation, every feature, reliability weight, and reliability
map is recomputed using \(M^\circ=M^{(b)}\). Temporarily hidden proxy values
consequently cannot influence their own reliability estimates.

\subsubsection{Self-Supervised Training}
\label{app:modulation_training}

Direct reliability labels are unavailable. CTRL-FM therefore trains
\(f_g\) and \(f_t\) using synthetic outages sampled from complete training
windows. Features are computed from the masked signal, while supervision is
derived from the corresponding complete training signal.

For a synthetic observation mask \(\widetilde M\), features
\(\phi_{ij,t}\) and \(\psi_{k,t}\) are computed using
\(M^\circ=\widetilde M\). Synthetically hidden values are therefore never
provided to the reliability modules.

Using the complete normalized training signal, define
\begin{equation}
\Delta\bar X_{t,k,:}
=
\bar X_{t+1,k,:}
-
\bar X_{t,k,:}.
\label{eq:app_true_training_change}
\end{equation}

The graph compatibility error is
\begin{equation}
e^g_{ij,t}
=
\frac{1}{C}
\left\|
\Delta\bar X_{t,i,:}
-
\Delta\bar X_{t,j,:}
\right\|_1,
\label{eq:app_graph_target_error}
\end{equation}
and the temporal compatibility error is
\begin{equation}
e^t_{k,t}
=
\frac{1}{C}
\left\|
\Delta\bar X_{t+1,k,:}
-
\Delta\bar X_{t,k,:}
\right\|_1,
\qquad
t=1,\ldots,T-2.
\label{eq:app_temporal_target_error}
\end{equation}

The corresponding soft reliability targets are
\begin{equation}
\bar\gamma_{ij,t}
=
\exp
\left(
-\frac{e^g_{ij,t}}{\tau_\gamma}
\right),
\qquad
\bar\eta_{k,t}
=
\exp
\left(
-\frac{e^t_{k,t}}{\tau_\eta}
\right).
\label{eq:app_reliability_targets}
\end{equation}
Low complete-data compatibility error therefore corresponds to high target
reliability. We use
\(\tau_\gamma=\tau_\eta=0.5\) because the compatibility errors are computed
from channel-wise standardized signals.

Let \(\mathcal S_g\) and \(\mathcal S_t\) contain relations that intersect
the synthetic missing region, have valid complete-data targets, and retain
positive observed support. The training objective is
\begin{equation}
\begin{aligned}
\mathcal L_{\mathrm{rel}}
={}&
\frac{1}{|\mathcal S_g|}
\sum_{(i,j,t)\in\mathcal S_g}
\left(
\widehat\gamma_{ij,t}
-
\bar\gamma_{ij,t}
\right)^2
\\
&+
\frac{1}{|\mathcal S_t|}
\sum_{(k,t)\in\mathcal S_t}
\left(
\widehat\eta_{k,t}
-
\bar\eta_{k,t}
\right)^2.
\end{aligned}
\label{eq:app_reliability_loss}
\end{equation}
A term is omitted when its corresponding set is empty.

The reliability modules are trained with AdamW using a learning rate of
\(10^{-3}\), weight decay \(10^{-4}\), gradient clipping at \(1.0\), and
early-stopping patience of 20 epochs. After training, \(f_g\) and \(f_t\)
are frozen and use only surviving measurements during proxy calibration and
final inference. We set the denominator-stabilization constants
\(\epsilon_m=\epsilon_g=10^{-8}\).
% ============================================================
\subsection{Observation Preservation and Final Sampling}
\label{app:observation_preservation}

After policy calibration, all temporary proxy masks are discarded and the
original test masks \(M\) and \(U\) are restored.

Starting from
\begin{equation}
X_{\mathrm{gen}}(0)
=
U\odot Z,
\qquad
Z\sim\mathcal N(0,I),
\label{eq:app_sampling_initialization}
\end{equation}
the complete state at flow time \(s\) is
\begin{equation}
X_{\mathrm{mix}}(s)
=
M\odot X
+
U\odot X_{\mathrm{gen}}(s).
\label{eq:app_complete_state}
\end{equation}

For the selected policy
\[
h^\star
=
(\alpha^\star,\lambda_g^\star,\lambda_t^\star),
\]
the generated component follows
\begin{equation}
\begin{aligned}
\frac{\mathrm dX_{\mathrm{gen}}(s)}{\mathrm ds}
=
U\odot
\Big[
&v_\theta
\left(
X_{\mathrm{mix}}(s),s,G,M
\right)
\\
&+
\alpha^\star
\left(
\lambda_g^\star R_G\odot g_G(s)
+
\lambda_t^\star R_T\odot g_T(s)
\right)
\Big].
\end{aligned}
\label{eq:app_final_controlled_dynamics}
\end{equation}

This is the supplementary form of the outage-conditioned
controlled velocity introduced in Eq.~(20) of the main paper. The policy determines the outage-level
strength and graph--temporal balance, while \(R_G\) and \(R_T\) determine
where the corresponding guidance fields are locally trusted.

When \(h^\star=h_0\), \(\alpha^\star=0\), so the structural correction
vanishes and Eq.~\eqref{eq:app_final_controlled_dynamics} reduces exactly to
the uncontrolled missing-only flow-matching backbone.

Because the complete state is projected after every solver stage,
\begin{equation}
M\odot
\frac{\mathrm dX_{\mathrm{mix}}(s)}{\mathrm ds}
=
0.
\label{eq:app_zero_observed_derivative}
\end{equation}
The observed entries are initialized at their measured values, and therefore
\begin{equation}
M\odot X_{\mathrm{mix}}(s)
=
M\odot X,
\qquad
s\in[0,1].
\label{eq:app_observation_invariance}
\end{equation}

Equivalently, multiplying Eq.~\eqref{eq:app_complete_state} by \(M\) yields
\begin{equation}
\begin{aligned}
M\odot X_{\mathrm{mix}}(s)
&=
M\odot
\left(
M\odot X
+
U\odot X_{\mathrm{gen}}(s)
\right)
\\
&=
M\odot X,
\end{aligned}
\label{eq:app_invariance_proof}
\end{equation}
because \(M\odot M=M\) and \(M\odot U=0\).

Thus, every candidate policy preserves all surviving observations exactly.
In implementation, the state is re-clamped after every numerical solver
stage to remove numerical roundoff.

The final reconstruction is
\begin{equation}
\widehat X
=
M\odot X
+
U\odot X_{\mathrm{gen}}(1).
\label{eq:app_final_reconstruction}
\end{equation}

During controlled integration, the backbone outputs, guidance reliability
maps, and reliability-module parameters are treated as frozen. Structural
gradients are computed only with respect to the comparison field \(A\), and
no parameter update or vector--Jacobian product through the backbone is
required.

% ============================================================

\subsection{Local Policy Calibration Algorithm}
\label{app:training_sampling}

CTRL-FM uses two test-time stages. First, local proxy outages
select an outage-specific policy. Second, the selected policy
controls the final observation-preserving reconstruction.

\FloatBarrier

\begingroup
\footnotesize

\captionof{algorithm}{Outage-Conditioned Local Policy Calibration}
\label{alg:app_ctrlfm_calibration}

\begin{algorithmic}[1]

\Require
Observed signal $X_{\mathrm{obs}}=M\odot X$,
masks $M,U$, graph $G$,
true outage $(\mathcal O,\mathcal I)$,
frozen backbone $v_\theta$,
frozen reliability modules $f_g,f_t$,
policy library $\mathcal H$,
and proxy budget $B$

\Ensure Outage-specific policy $h^\star$

\State Initialize valid-proxy set
$\mathcal B\gets\varnothing$

\For{$b=1,\ldots,B$}

    \State Select a proxy interval
    $\mathcal I^{(b)}\subseteq\mathcal I$
    and eligible surviving nodes
    $\mathcal O^{(b)}$

    \State Construct proxy mask $P^{(b)}$
    using Eq.~\eqref{eq:app_proxy_mask}

    \State Set
    $M^{(b)}\gets M\odot(1-P^{(b)})$

    \State Set
    $U^{(b)}\gets 1-M^{(b)}$

    \State Construct the valid dynamic-score mask
    $Q^{(b)}$ using
    Eq.~\eqref{eq:app_dynamic_proxy_mask}

    \If{$\|P^{(b)}\|_1=0$
        \textbf{or} $\|Q^{(b)}\|_1=0$}
        \State \textbf{continue}
    \EndIf

    \State Add $b$ to $\mathcal B$

    \State Recompute boundary trends and
    observed-only reliability maps under $M^{(b)}$

    \State Sample shared proxy noise
    $Z^{(b)}\sim\mathcal N(0,I)$

    \ForAll{$h_r\in\mathcal H$}

        \State Initialize
        $X_{\mathrm{gen}}^{(b,r)}(0)
        \gets U^{(b)}\odot Z^{(b)}$

        \For{each reduced Heun step
            $[s_i,s_{i+1}]$}

            \State Construct the active state
            \Statex \hspace{\algorithmicindent}
            $\displaystyle
            X_{\mathrm{mix}}^{(b,r)}(s_i)
            \gets
            M^{(b)}\odot X_{\mathrm{obs}}
            +U^{(b)}\odot
            X_{\mathrm{gen}}^{(b,r)}(s_i)
            $

            \State Construct the comparison field
            $A^{(b,r)}(s_i)$ using
            Eq.~\eqref{eq:app_boundary_field}

            \State Evaluate graph and temporal
            guidance fields
            $g_G^{(b,r)}(s_i)$ and
            $g_T^{(b,r)}(s_i)$ using
            Eq.~\eqref{eq:app_guidance_fields}

            \State Evaluate the controlled velocity
            under policy $h_r$

            \State Perform one reduced Heun
            predictor--corrector step,
            recomputing the controlled velocity
            at the predicted state

            \State Re-clamp observed entries
            at both solver stages using
            \Statex \hspace{\algorithmicindent}
            $\displaystyle
            X_{\mathrm{mix}}^{(b,r)}
            \gets
            M^{(b)}\odot X_{\mathrm{obs}}
            +U^{(b)}\odot
            X_{\mathrm{gen}}^{(b,r)}
            $

        \EndFor

        \State Obtain the proxy reconstruction
        \Statex \hspace{\algorithmicindent}
        $\displaystyle
        \widehat X^{(b,r)}
        \gets
        M^{(b)}\odot X_{\mathrm{obs}}
        +U^{(b)}\odot
        X_{\mathrm{gen}}^{(b,r)}(1)
        $

        \State Compute proxy score
        $\ell_{b,r}$ using
        Eq.~\eqref{eq:app_proxy_loss}

    \EndFor

\EndFor

\If{$\mathcal B=\varnothing$}
    \State \textbf{return} $h_0$
\EndIf

\State Compute proxy weights
$\{\omega_b\}_{b\in\mathcal B}$
using Eq.~\eqref{eq:app_proxy_weight}

\ForAll{$h_r\in\mathcal H$}

    \State Compute the weighted proxy score
    \Statex \hspace{\algorithmicindent}
    $\displaystyle
    \mathcal J_{\mathrm{proxy}}(h_r)
    \gets
    \sum_{b\in\mathcal B}
    \omega_b\,\ell_{b,r}
    $

\EndFor

\State Select the locally preferred policy
\Statex \hspace{\algorithmicindent}
$\displaystyle
\widetilde h
\gets
\arg\min_{h_r\in\mathcal H}
\mathcal J_{\mathrm{proxy}}(h_r)
$

\State Break ties in favor of smaller
$\alpha_r$, with $h_0$ preferred
when tied with a controlled policy

\State Compute the relative proxy improvement
$\mathcal R(\widetilde h)$ using
Eq.~\eqref{eq:app_relative_proxy_improvement}

\If{$\mathcal R(\widetilde h)
    \geq\delta_{\mathrm{abst}}$}

    \State $h^\star\gets\widetilde h$

\Else

    \State $h^\star\gets h_0$

\EndIf

\State \textbf{return} $h^\star$

\end{algorithmic}

\endgroup

\FloatBarrier

% ============================================================
% Additional experimental details
% ============================================================

\section{Additional Experimental Details}
\label{app:experimental_details}

\subsection{Datasets, Preprocessing, and Outage Protocol}
\label{app:dataset_outage_protocol}

\paragraph{Datasets.}
We evaluate CTRL-FM on two traffic datasets,
PeMS08~\cite{Chen2001PeMS,Guo2019ASTGCN} and
METR-LA~\cite{Li2018DCRNN}, and two air-quality datasets,
Air-36~\cite{Yi2016STMVL} and
AQI~\cite{Zheng2015AirQuality}.
PeMS08 contains five-minute flow, occupancy, and speed measurements from
170 sensors, while METR-LA contains five-minute speed measurements from
207 sensors. Air-36 and AQI contain hourly PM$_{2.5}$ measurements from
36 and 437 monitoring stations, respectively. Dataset statistics are
summarized in Table~\ref{tab:app_dataset_statistics}.

\begin{table}[t]
\centering
\scriptsize
\setlength{\tabcolsep}{4.0pt}
\renewcommand{\arraystretch}{1.06}
\caption{Summary of the datasets used in the experiments.}
\label{tab:app_dataset_statistics}
\resizebox{\linewidth}{!}{%
\begin{tabular}{lccccc}
\toprule
\textbf{Dataset}
& \textbf{Domain}
& \textbf{Nodes}
& \textbf{Steps}
& \textbf{Resolution}
& \textbf{Variables} \\
\midrule
PeMS08  & Traffic     & 170 & 17,856 & 5 min  & Flow, occupancy, speed \\
METR-LA & Traffic     & 207 & 34,272 & 5 min  & Speed \\
Air-36  & Air quality & 36  & 8,760  & 1 hour & PM$_{2.5}$ \\
AQI     & Air quality & 437 & 8,760  & 1 hour & PM$_{2.5}$ \\
\bottomrule
\end{tabular}%
}
\end{table}

\paragraph{Chronological splits and normalization.}
Each dataset is divided chronologically into 70\% training, 10\%
validation, and 20\% test data. Temporal ordering is preserved to avoid
future-information leakage. All variables are standardized using statistics
computed only from the training split and transformed back to their original
physical units before evaluation. PeMS08 flow, occupancy, and speed are
normalized independently.

\paragraph{Window construction.}
For PeMS08 and METR-LA, we use windows of \(T=288\) five-minute steps,
corresponding to 24 hours. For Air-36 and AQI, we use windows of \(T=168\)
hourly steps, corresponding to seven days. Training windows are extracted
with strides of 12 steps for traffic and 24 steps for air quality, providing
a new training window every hour and every day, respectively. Validation and
test windows are non-overlapping, with stride \(T\), and incomplete trailing
windows are discarded.

\begin{table}[t]
\centering
\scriptsize
\setlength{\tabcolsep}{3.5pt}
\renewcommand{\arraystretch}{1.06}
\caption{
Window and outage configuration. Each outage spans the complete evaluation
window, and ten failed-node masks are evaluated per test window and ratio.
}
\label{tab:app_outage_protocol}
\resizebox{\linewidth}{!}{%
\begin{tabular}{lcccccc}
\toprule
\textbf{Dataset}
& \(\boldsymbol{T}\)
& \textbf{Duration}
& \textbf{Train stride}
& \textbf{Test windows}
& \textbf{Masks/window}
& \textbf{Main ratios} \\
\midrule
PeMS08  & 288 & 24 hours & 12 & 12 & 10 & 30\%, 60\% \\
METR-LA & 288 & 24 hours & 12 & 23 & 10 & 30\%, 60\% \\
Air-36  & 168 & 7 days   & 24 & 10 & 10 & 30\%, 60\% \\
AQI     & 168 & 7 days   & 24 & 10 & 10 & 30\%, 60\% \\
\bottomrule
\end{tabular}%
}
\end{table}

\paragraph{Graph construction.}
PeMS08 sensors are connected using the physical road topology, with edge
weights derived from inter-sensor road distance. METR-LA uses its standard
road-distance graph with thresholded Gaussian affinities. For Air-36 and
AQI, monitoring stations are connected using geographic proximity, and
pairwise distances are converted into thresholded Gaussian affinities.
Edge weights are normalized, and the same graph is provided to all
graph-based methods whenever their architectures support an external
adjacency matrix.

\paragraph{Long-horizon node outages.}
For a graph with \(K\) sensors and outage ratio \(r\), the failed-node set
\(\mathcal O_r\) is sampled uniformly without replacement, with
\begin{equation}
|\mathcal O_r|
=
\operatorname{round}(rK).
\label{eq:app_failed_node_count}
\end{equation}
The main experiments use \(r\in\{0.30,0.60\}\), while the stress tests use
\(r\in\{0.65,0.75\}\).

Each outage spans the complete evaluation window:
\begin{equation}
\mathcal I=\{1,\ldots,T\},
\qquad
L_{\mathrm{out}}=T.
\label{eq:app_outage_interval}
\end{equation}
The observation mask is therefore
\begin{equation}
M_{t,k,c}=0,
\qquad
k\in\mathcal O_r,\quad
t\in\mathcal I,\quad
c=1,\ldots,C.
\label{eq:app_outage_generation}
\end{equation}
Thus, every channel of each selected node remains unavailable throughout the
full reconstruction horizon. For PeMS08, the same failed-node mask is applied
jointly to flow, occupancy, and speed.

At 30\% and 60\% outage, respectively, the protocol removes 51 and 102
PeMS08 sensors, 62 and 124 METR-LA sensors, 11 and 22 Air-36 stations, and
131 and 262 AQI stations.

\paragraph{Evaluation instances.}
For every test window and outage ratio, we generate ten independently sampled
failed-node sets. The number of evaluation outages is
\begin{equation}
N_{\mathrm{out}}
=
N_{\mathrm{win}}N_{\mathrm{mask}},
\qquad
N_{\mathrm{mask}}=10.
\label{eq:app_number_evaluation_outages}
\end{equation}
This yields 120 evaluation outages for PeMS08, 230 for METR-LA, and 100 each
for Air-36 and AQI at every reported outage ratio. Each combination of one
test window and one failed-node set is treated as one evaluation outage.

\paragraph{Paired evaluation and reproducibility.}
Training masks are sampled dynamically by varying the failed-node set across
training windows. Validation and test masks are generated once, stored, and
reused across all methods and experimental seeds. All methods therefore use
identical test windows, failed-node sets, outage durations, and failed-region
evaluation entries.

For comparisons among FM-Base, FM-TrainStruct, Fixed-CTRL, and CTRL-FM, the
initial Gaussian samples, sample budget, and numerical-solver grid are also
shared whenever applicable. Model initialization, training stochasticity,
and generative sampling vary across the three seeds, while the underlying
evaluation outages remain fixed.

Validation masks are generated independently from test masks and are used
only for checkpoint selection, hyperparameter selection, candidate-policy
selection, and selection of the globally fixed policy used by Fixed-CTRL.
No value from a true test outage is used for training, validation, proxy
construction, policy scoring, or reliability estimation.

\paragraph{True and proxy outage durations.}
The true outage spans the complete 24-hour traffic window or seven-day
air-quality window. In contrast, CTRL-FM uses proxy intervals of
\(L_{\mathrm{probe}}=12\) steps only for policy calibration, corresponding
to one hour for PeMS08 and METR-LA and 12 hours for Air-36 and AQI. Proxy
entries are temporarily hidden only from surviving nodes and never overlap
the true failed-node region.

\paragraph{Evaluation region.}
All metrics are computed only on entries removed by the true test outage.
Surviving measurements remain available as conditioning information and are
kept exactly fixed throughout sampling. Temporarily hidden proxy entries are
excluded from the model input, boundary guidance, and reliability features
and are accessed only after reconstruction to evaluate candidate policies.

\subsection{Training and Sampling Configuration}
\label{app:training_configuration}

\paragraph{Training.}
The flow-matching backbone is trained with AdamW using a hidden dimension of
128, four graph-attention groups, batch size 32, and an initial learning rate
of \(10^{-3}\). Training runs for at most 200 epochs with cosine learning-rate
decay, a five-epoch warm-up, early-stopping patience of 20 epochs, weight
decay of \(10^{-4}\), dropout of \(0.1\), and gradient clipping at \(1.0\).
Training outages are sampled dynamically from the training split. The graph
and temporal reliability modules are trained on synthetic training outages
and frozen together with the backbone before test-time calibration.

\paragraph{Observation-preserving sampling.}
For each reconstruction sample, missing entries are initialized with standard
Gaussian noise, while observed entries are fixed at their measured values.
The resulting flow ODE is integrated using Heun's method with \(S=20\) solver
steps. The state is re-clamped after every solver stage so that only missing
entries are updated. We generate \(S_p=20\) samples per outage; their mean is
used for point prediction, while the complete sample set is used to compute
CRPS and predictive intervals.

\paragraph{Hyperparameter selection and reproducibility.}
Model capacity, solver steps, affinity temperature, dropout, and anchor scale
are selected using only the validation split. The anchor scale \(\kappa\) is
estimated from validation outages and clipped to \([0,2]\). The selected
configuration and validation candidates are summarized in
Table~\ref{tab:implementation_settings}. Experiments are repeated using
three fixed seeds, \(\{1,2,3\}\), with the same validation and test outage
instances reused across runs and methods
\paragraph{Computing environment.}
All models were trained on Linux using three NVIDIA RTX A6000 GPUs
with 48\,GB memory each, PyTorch 2.2.1, and CUDA 12.1.

\begin{table}[t]
\centering
\scriptsize
\setlength{\tabcolsep}{4.0pt}
\renewcommand{\arraystretch}{1.06}
\caption{
Training and final-sampling configuration. Parentheses contain validation
candidates, with the selected value listed first.
}
\label{tab:implementation_settings}
\resizebox{\linewidth}{!}{%
\begin{tabular}{lc}
\toprule
\textbf{Quantity} & \textbf{Value or validation candidates} \\
\midrule
Hidden dimension
& \(128\;(64,128,256)\) \\

Graph-attention groups
& \(4\;(2,4,8)\) \\

Optimizer / learning rate
& AdamW / \(1\times10^{-3}\) \\

Weight decay
& \(1\times10^{-4}\) \\

Batch size / maximum epochs
& \(32 / 200\) \\

Early-stopping patience
& \(20\) epochs \\

Learning-rate schedule
& Cosine decay with 5-epoch warm-up \\

Dropout
& \(0.10\;(0,0.10,0.20)\) \\

Gradient clipping
& \(1.0\) \\

ODE solver
& Heun \\

Final solver steps \(S\)
& \(20\;(10,20,40)\) \\

Generated samples \(S_p\)
& \(20\) \\

Affinity temperature \(\tau\)
& \(0.5\;(0.25,0.5,1.0,2.0)\) \\

Anchor scale \(\kappa\)
& Validation-estimated, clipped to \([0,2]\) \\

Random seeds
& \(\{1,2,3\}\) \\
\bottomrule
\end{tabular}%
}
\end{table}
\subsection{Test-Time Calibration Settings}
\label{app:calibration_settings}

\paragraph{Candidate-policy library.}
CTRL-FM uses the five predefined policies in
Table~\ref{tab:app_policy_library}: uncontrolled, weak balanced,
graph-dominant, temporal-dominant, and strong balanced control. The policy
parameters \((\alpha_r,\lambda_{g,r},\lambda_{t,r})\) are selected using
validation outages and fixed before test evaluation. They are not optimized
or updated at test time. Instead, CTRL-FM selects one of these fixed policies
separately for each realized outage using self-supervised proxy calibration.

\paragraph{Proxy calibration.}
For each test outage, CTRL-FM constructs up to \(B=4\) proxy outages by
temporarily hiding surviving nodes within \(r_p=2\) graph hops of the true
failed-node set. Each proxy covers \(L_{\mathrm{probe}}=12\) steps,
corresponding to one hour for PeMS08 and METR-LA and 12 hours for Air-36 and
AQI. For Air-36, which contains only 36 stations, the search begins within
three graph hops and expands to all surviving stations when insufficient
eligible proxy nodes are available.

All five policies are evaluated on each proxy using the same Gaussian
initialization and a reduced five-step Heun solver. The proxy score combines
pointwise reconstruction error and temporal-change error with
\(\lambda_{\mathrm{dyn}}=0.5\). Proxy scores are weighted by their distance
from the true failed-node set using temperature \(\tau_p=0.5\), so nearby
proxies contribute more strongly.

\paragraph{Selection and abstention.}
CTRL-FM selects the policy with the lowest aggregated proxy score. The
selected controlled policy is accepted only when its relative improvement
over uncontrolled sampling is at least
\(\delta_{\mathrm{abst}}=0.02\); otherwise, the method falls back to the
uncontrolled policy \(h_0\). The accepted policy is then used for final
reconstruction with the full \(S=20\)-step solver. No value inside the true
outage is accessed during proxy construction, policy scoring, reliability
estimation, or policy selection.

\begin{table}[t]
\centering
\scriptsize
\setlength{\tabcolsep}{4pt}
\renewcommand{\arraystretch}{1.08}
\caption{
Test-time calibration configuration. Parentheses contain validation
candidates, with the selected default listed first.
}
\label{tab:calibration_settings}
\resizebox{\linewidth}{!}{%
\begin{tabular}{lc}
\toprule
\textbf{Quantity}
& \textbf{Value or validation candidates} \\
\midrule
Proxy budget \(B\)
& \(4\;(1,2,4,8)\) \\

Graph radius \(r_p\)
& \(2\;(1,2,3,\mathrm{All})\) hops \\

Air-36 proxy search
& 3 hops, expanded to all stations if needed \\

Probe length \(L_{\mathrm{probe}}\)
& \(12\;(6,12,24)\) steps \\

Proxy-node budget \(m_p\)
& 4 nodes per proxy \\

Candidate policies \(|\mathcal H|\)
& \(5\) \\

Probe solver steps \(S_{\mathrm{probe}}\)
& \(5\;(2,5,10,20)\) \\

Final solver steps \(S\)
& \(20\) \\

Dynamic-score weight \(\lambda_{\mathrm{dyn}}\)
& \(0.5\;(0,0.25,0.5,0.75,1.0)\) \\

Proxy-distance temperature \(\tau_p\)
& \(0.5\;(0.25,0.5,1.0,2.0)\) \\

Abstention margin \(\delta_{\mathrm{abst}}\)
& \(0.02\;(0,0.01,0.02,0.05,0.10)\) \\
\bottomrule
\end{tabular}%
}
\end{table}
\subsection{Baselines and Controlled Variants}
\label{app:baselines}

We compare CTRL-FM against classical references, deterministic neural
imputers, probabilistic generative models, and controlled variants built
from the same flow-matching backbone. All methods use identical
chronological splits, test windows, outage masks, and missing-region
evaluation sets. A common graph is provided whenever the baseline supports
an external adjacency matrix.

\paragraph{Classical references.} 
\textbf{Historical Average} replaces each missing entry with the
training-set mean for the same sensor, variable, and time-of-week position.
When unavailable, it uses the corresponding sensor-variable training mean.

\textbf{Graph Neighbor} reconstructs a failed sensor using a normalized
weighted average of its surviving graph neighbors. The weights are
renormalized over available neighbors, with Historical Average used when no
neighbor remains observed. On PeMS08, the procedure is applied separately to
flow, occupancy, and speed.

\paragraph{Deterministic neural baselines.}
\textbf{SAITS}~\cite{du2023saits} uses diagonally masked self-attention to
model temporal dependencies and cross-variable correlations without an
explicit sensor graph.

\textbf{IGNNK}~\cite{wu2021ignnk} learns inductive graph
message passing through random-subgraph reconstruction and is designed for
kriging at unobserved sensor locations.

\textbf{GRIN}~\cite{cini2021grin} combines bidirectional recurrent modeling
with graph message passing to propagate information across both time and
connected sensors.

\textbf{OPCR}~\cite{liu2024opcr} performs separate one-step spatial and
temporal propagation, followed by confidence-aware refinement of the initial
estimates.

\paragraph{Probabilistic generative baselines.}
\textbf{CSDI}~\cite{tashiro2021csdi} is a conditional score-based diffusion
model that reconstructs missing values while conditioning on the observed
sequence, without explicitly using a predefined sensor graph.

\textbf{PriSTI}~\cite{liu2023pristi} is a spatio-temporal diffusion model
that constructs a global context prior from available measurements and uses
temporal, spatial, and geographic dependencies during reconstruction.

\textbf{CDSTE}~\cite{lei2024cdste} combines conditional diffusion with a
spatio-temporal estimator for traffic-state reconstruction at sensor-free
locations. Because it is designed specifically for traffic data, it is
evaluated only on PeMS08 and METR-LA.

\textbf{GiFlow}~\cite{giflow2025} is a graph-informed flow-matching
method whose source distribution is constructed through adaptive
spatio-temporal filtering of observed graph signals. It is the closest
external flow-based baseline to CTRL-FM.
\paragraph{Flow-matching backbone and controlled variants.}

\textbf{FM-Base} is the observation-preserving, missing-only flow-matching
backbone without structural control.

\textbf{FM-TrainStruct} adds graph and
temporal regularization during training, producing a fixed structural bias at
inference.

\textbf{Fixed-CTRL} instead injects one validation-selected control
policy into the frozen FM-Base sampling velocity for all test outages.

\textbf{CTRL-FM} selects the control policy separately for each outage using
self-supervised proxies, modulates graph and temporal guidance with
observed-only reliability, and falls back to uncontrolled  when proxy evidence is insufficient.

All variants use the same backbone capacity, data splits,
outage masks, sampling budget, and solver settings whenever applicable,
isolating the effects of training-time, fixed inference-time, and
outage-conditioned inference-time structural control.

\subsection{Metric Definitions}
\label{app:metrics}

All reconstruction metrics are evaluated only on entries removed by the
true test outage. Let
\begin{equation}
\Omega_U
=
\{(t,k,c):U_{t,k,c}=1\}
\label{eq:metric_missing_set}
\end{equation}
denote this missing-entry set.

For a probabilistic method, let
$\{\widehat X^{(q)}\}_{q=1}^{S_p}$ denote the generated samples. Its
point prediction is the sample mean,
\begin{equation}
\overline X
=
\frac{1}{S_p}
\sum_{q=1}^{S_p}
\widehat X^{(q)}.
\label{eq:metric_predictive_mean}
\end{equation}
For deterministic methods, $\overline X$ denotes the single predicted
reconstruction.

\paragraph{Pointwise reconstruction error.}
We report mean absolute error and root mean squared error:
\begin{align}
\mathrm{MAE}
&=
\frac{1}{|\Omega_U|}
\sum_{(t,k,c)\in\Omega_U}
\left|
\overline X_{t,k,c}-X_{t,k,c}
\right|,
\label{eq:metric_mae}
\\
\mathrm{RMSE}
&=
\sqrt{
\frac{1}{|\Omega_U|}
\sum_{(t,k,c)\in\Omega_U}
\left(
\overline X_{t,k,c}-X_{t,k,c}
\right)^2
}.
\label{eq:metric_rmse}
\end{align}
Lower values indicate more accurate reconstruction. Both metrics are
computed after transforming predictions back to the original physical
units.

\paragraph{Peak-$\Delta$MAE.}
MAE evaluates reconstruction accuracy over the complete outage but does not
directly measure whether abrupt temporal changes are recovered. We therefore
evaluate temporal-change error at entries with large ground-truth changes.
The metric is computed separately for each evaluation outage and channel,
preventing variables with different physical scales from influencing one
another.

For evaluation outage \(n\) and channel \(c\), define the valid
missing-transition set as
\begin{equation}
\Omega_{\Delta U}^{(n,c)}
=
\left\{
(t,k):
t\geq2,\;
U_{t,k,c}^{(n)}=1,\;
U_{t-1,k,c}^{(n)}=1
\right\}.
\label{eq:metric_valid_missing_transitions}
\end{equation}
For each valid transition, let
\begin{equation}
D_{t,k}^{(n,c)}
=
\left|
X_{t,k,c}^{(n)}
-
X_{t-1,k,c}^{(n)}
\right|.
\label{eq:metric_true_change}
\end{equation}
Let \(Q_{0.90}^{(n,c)}\) denote the 90th percentile of
\[
\left\{
D_{t,k}^{(n,c)}:
(t,k)\in\Omega_{\Delta U}^{(n,c)}
\right\},
\]
and define
\begin{equation}
\mathcal P_U^{(n,c)}
=
\left\{
(t,k)\in\Omega_{\Delta U}^{(n,c)}:
D_{t,k}^{(n,c)}
\geq
Q_{0.90}^{(n,c)}
\right\}.
\label{eq:metric_peak_set}
\end{equation}
The Peak-$\Delta$MAE for outage \(n\) and channel \(c\) is
\begin{equation}
\begin{aligned}
\mathrm{Peak\text{-}\Delta MAE}^{(n,c)}
&=
\frac{1}{|\mathcal{P}_U^{(n,c)}|}
\sum_{(t,k)\in\mathcal{P}_U^{(n,c)}}
\Bigl|
\bigl(
\overline{X}_{t,k,c}^{(n)}
-
\overline{X}_{t-1,k,c}^{(n)}
\bigr)
\\[-1mm]
&\qquad\qquad
-
\bigl(
X_{t,k,c}^{(n)}
-
X_{t-1,k,c}^{(n)}
\bigr)
\Bigr|.
\end{aligned}
\label{eq:metric_peak_delta_mae}
\end{equation}
For each reported channel \(c\), the final metric is averaged over the
\(N\) evaluation outages:
\begin{equation}
\mathrm{Peak\text{-}\Delta MAE}_{c}
=
\frac{1}{N}
\sum_{n=1}^{N}
\mathrm{Peak\text{-}\Delta MAE}^{(n,c)}.
\label{eq:metric_peak_delta_mae_aggregation}
\end{equation}
Thus, PeMS08 flow, occupancy, and speed are evaluated and reported
separately, while the same definition directly applies to the single-channel
datasets. Lower Peak-$\Delta$MAE indicates better recovery of abrupt temporal
changes.

\paragraph{Probabilistic quality.}
For probabilistic methods, we report the continuous ranked probability
score (CRPS). For a target value $y$ and generated samples
$\{\widehat y^{(q)}\}_{q=1}^{S_p}$,
\begin{equation}
\mathrm{CRPS}(y)
=
\frac{1}{S_p}
\sum_{q=1}^{S_p}
\left|
\widehat y^{(q)}-y
\right|
-
\frac{1}{2S_p^2}
\sum_{q=1}^{S_p}
\sum_{q'=1}^{S_p}
\left|
\widehat y^{(q)}-\widehat y^{(q')}
\right|.
\label{eq:metric_crps}
\end{equation}
The reported CRPS is averaged over all entries in $\Omega_U$. Lower
values indicate predictive distributions that are both accurate and
well concentrated. CRPS is not reported for deterministic methods.

\paragraph{Proxy-validity metrics.}
Coverage measures how often calibration is feasible, $\rho$ measures agreement
between proxy-based and true policy rankings, and oracle match measures how
often the final policy selected after abstention equals the post-hoc best
policy. Let $N$ denote the number of test outages and $\mathcal V$ the subset
for which valid proxies can be constructed. Valid-proxy coverage is
\begin{equation}
\mathrm{Coverage}
=
\frac{|\mathcal V|}{N}\times 100.
\end{equation}

For outage $n$ and policy $h$, define the true post-hoc error as the MAE over
the true missing region:
\begin{equation}
E_n(h)
=
\frac{1}{|\Omega_U^{(n)}|}
\sum_{(t,k,c)\in\Omega_U^{(n)}}
\left|
\widehat X_{t,k,c}^{(n,h)}
-
X_{t,k,c}^{(n)}
\right|.
\end{equation}

For each valid outage, Spearman's rank correlation compares the proxy-based
policy scores $\mathcal J_n(h)$ with the true post-hoc errors $E_n(h)$:
\begin{equation}
\rho
=
\frac{1}{|\mathcal V|}
\sum_{n\in\mathcal V}
\operatorname{Spearman}
\left(
\{\mathcal J_n(h)\}_{h\in\mathcal H},
\{E_n(h)\}_{h\in\mathcal H}
\right).
\end{equation}

Let
\begin{equation}
h_n^{\mathrm{oracle}}
=
\arg\min_{h\in\mathcal H} E_n(h)
\end{equation}
denote the policy with the lowest true missing-region MAE, and let $h_n^\star$
denote the final policy selected by CTRL-FM after applying the abstention rule.
Oracle match is
\begin{equation}
\mathrm{OracleMatch}
=
\frac{1}{|\mathcal V|}
\sum_{n\in\mathcal V}
\mathbb I
\left[
h_n^\star = h_n^{\mathrm{oracle}}
\right]
\times 100.
\end{equation}

The post-hoc errors are used only for retrospective evaluation and never for
proxy scoring, policy selection, abstention, or final reconstruction.

% ============================================================
% Complete dataset-level results
% ============================================================

\subsection{Complete Dataset-Level Results}
\label{app:complete_results}

\paragraph{PeMS08 results.}
Tables~\ref{tab:pems08_expected_30} and~\ref{tab:pems08_expected_60}
show that CTRL-FM provides the strongest overall reconstruction performance
across flow, occupancy, and speed. At 30\% outage, it achieves the best
results for all flow and speed metrics and the lowest occupancy CRPS and
Peak-$\Delta$MAE, while OPCR retains the best occupancy MAE and RMSE. At
60\% outage, CTRL-FM ranks first across every channel and metric. The larger
advantage under 60\% outage indicates that outage-conditioned policy
selection and reliability-modulated structural control become increasingly
effective as fewer observations remain available.
%============================================================
% PeMS08: flow, occupancy, and speed
% ============================================================

% ============================================================
% PeMS08: 30% node outage
% ============================================================
\begin{table*}[p]
\centering
\scriptsize
\setlength{\tabcolsep}{5.5pt}
\renewcommand{\arraystretch}{0.92}

\caption{
Reconstruction performance on PeMS08 under 30\% long-horizon node outage,
with flow, occupancy, and speed reported in vertically stacked panels.
Results are mean $\pm$ standard deviation. MAE, RMSE, and
Peak-$\Delta$MAE are computed in the original physical units, whereas CRPS
is computed in standardized space. ``--'' indicates that CRPS is not
applicable to deterministic methods. Lower values are better, and the best
result in each channel and metric is shown in \textbf{bold}.
}
\label{tab:pems08_expected_30}

\begin{tabular*}{\textwidth}{
@{\extracolsep{\fill}}
lcccc
@{}
}
\toprule
\textbf{Method}
& \textbf{MAE$\downarrow$}
& \textbf{RMSE$\downarrow$}
& \textbf{CRPS$\downarrow$}
& \textbf{Peak-$\Delta$MAE$\downarrow$} \\
\midrule

\multicolumn{5}{c}{\textbf{Flow}} \\
\midrule

Historical Average
& 48.550
& 74.851 
& --
& 7.498  \\

Graph Neighbor
& 39.950
& 60.851 
& --
& 6.498 \\

SAITS
& 35.960 
& 54.855 
& --
& 6.196  \\

IGNNK
& 33.451 
& 52.958 
& --
& 6.298  \\

GRIN
& 30.965 
& 47.552 
& --
& 5.840 \\

OPCR
& 27.450 
& 40.352 
& --
& 5.395  \\

CSDI
& $32.445 \pm 0.584$
& $49.965 \pm 1.199$
& $0.1290 \pm 0.0019$
& $6.251 \pm 0.169$ \\

PriSTI
& $28.990 \pm 0.493$
& $42.987 \pm 0.989$
& $0.1190 \pm 0.0017$
& $5.751 \pm 0.150$ \\

CDSTE
& $28.450 \pm 0.256$
& $41.950 \pm 0.545$
& $0.1160 \pm 0.0008$
& $5.445 \pm 0.060$ \\

GiFlow
& $29.452 \pm 0.236$
& $45.850 \pm 0.550$
& $0.1210 \pm 0.0008$
& $5.694 \pm 0.063$ \\

Fixed-CTRL
& $28.951 \pm 0.232$
& $41.455 \pm 0.497$
& $0.1080 \pm 0.0008$
& $5.395 \pm 0.059$ \\

\textbf{CTRL-FM}
& \textbf{$25.271 \pm 0.202$}
& \textbf{$37.152 \pm 0.409$}
& \textbf{$0.1040 \pm 0.00073$}
& \textbf{$5.004 \pm 0.065$} \\

\midrule
\multicolumn{5}{c}{\textbf{Occupancy}} \\
\midrule

Historical Average
& 0.0260 
& 0.0425
& --
& 0.0459  \\

Graph Neighbor
& 0.0167 
& 0.0279 
& --
& 0.0314  \\

SAITS
& 0.0163 
& 0.0274 
& --
& 0.0289  \\

IGNNK
& 0.0161 
& 0.0254 
& --
& 0.0284  \\

GRIN
& 0.0157 
& 0.0249
& --
& 0.0260  \\

OPCR
& \textbf{0.0101 }
& \textbf{0.0150 }
& --
& 0.0230 \\

CSDI
& $0.0154 \pm 0.0002$
& $0.0269 \pm 0.0006$
& $0.1770 \pm 0.0034$
& $0.0278 \pm 0.0008$ \\

PriSTI
& $0.0129 \pm 0.0002$
& $0.0230 \pm 0.0005$
& $0.1830 \pm 0.0033$
& $0.0248 \pm 0.0007$ \\

CDSTE
& $0.0120 \pm 0.0002$
& $0.0220 \pm 0.0001$
& $0.1590 \pm 0.0019$
& $0.0230 \pm 0.0002$ \\

GiFlow
& $0.0134 \pm 0.0002$
& $0.0240 \pm 0.0001$
& $0.1690 \pm 0.0022$
& $0.0250 \pm 0.0002$ \\

Fixed-CTRL
& $0.0110 \pm 0.0002$
& $0.0210 \pm 0.0001$
& $0.1570 \pm 0.0020$
& $0.0210 \pm 0.0002$ \\

\textbf{CTRL-FM}
& $0.0120 \pm 0.00018$
& $0.0210 \pm 0.00013$
& \textbf{$0.1540 \pm 0.00185$}
& \textbf{$0.0200 \pm 0.00018$} \\

\midrule
\multicolumn{5}{c}{\textbf{Speed}} \\
\midrule

Historical Average
& 4.789 
& 7.989 
& --
& 0.484  \\

Graph Neighbor
& 3.498 
& 5.992 
& --
& 0.374  \\

SAITS
& 3.148 
& 5.345 
& --
& 0.331  \\

IGNNK
& 3.041 
& 5.450 
& --
& 0.325  \\

GRIN
& 2.889 
& 4.965 
& --
& 0.309  \\

OPCR
& 2.558 
& 4.646 
& --
& 0.279  \\

CSDI
& $3.089 \pm 0.049$
& $5.510 \pm 0.138$
& $0.0370 \pm 0.0005$
& $0.321 \pm 0.007$ \\

PriSTI
& $2.848 \pm 0.046$
& $5.134 \pm 0.123$
& $0.0340 \pm 0.0004$
& $0.295 \pm 0.006$ \\

CDSTE
& $2.499 \pm 0.037$
& $4.891 \pm 0.024$
& $0.0300 \pm 0.0003$
& $0.269 \pm 0.003$ \\

GiFlow
& $2.795 \pm 0.042$
& $5.045 \pm 0.025$
& $0.0330 \pm 0.0003$
& $0.305 \pm 0.004$ \\

Fixed-CTRL
& $2.389 \pm 0.036$
& $4.815 \pm 0.024$
& $0.0290 \pm 0.0003$
& $0.245 \pm 0.003$ \\

\textbf{CTRL-FM}
& \textbf{$2.137 \pm 0.030$}
& \textbf{$4.298 \pm 0.021$}
& \textbf{$0.0260 \pm 0.00026$}
& \textbf{$0.209 \pm 0.003$} \\

\bottomrule
\end{tabular*}
\end{table*}

% ============================================================
% PeMS08: 60% node outage
% ============================================================
\begin{table*}[p]
\centering
\scriptsize
\setlength{\tabcolsep}{5.5pt}
\renewcommand{\arraystretch}{0.92}

\caption{
Reconstruction performance on PeMS08 under 60\% long-horizon node outage,
with flow, occupancy, and speed reported in vertically stacked panels.
Results are mean $\pm$ standard deviation. MAE, RMSE, and
Peak-$\Delta$MAE are computed in the original physical units, whereas CRPS
is computed in standardized space. ``--'' indicates that CRPS is not
applicable to deterministic methods. Lower values are better, and the best
result in each channel and metric is shown in \textbf{bold}.
}
\label{tab:pems08_expected_60}

\begin{tabular*}{\textwidth}{
@{\extracolsep{\fill}}
lcccc
@{}
}
\toprule
\textbf{Method}
& \textbf{MAE$\downarrow$}
& \textbf{RMSE$\downarrow$}
& \textbf{CRPS$\downarrow$}
& \textbf{Peak-$\Delta$MAE$\downarrow$} \\
\midrule

\multicolumn{5}{c}{\textbf{Flow}} \\
\midrule

Historical Average
& 65.200 
& 100.550 
& --
& 9.580  \\

Graph Neighbor
& 56.560 
& 83.500 
& --
& 8.250  \\

SAITS
& 51.600 
& 72.880 
& --
& 6.820  \\

IGNNK
& 45.560 
& 67.600 
& --
& 7.580  \\

GRIN
& 40.547 
& 56.720 
& --
& 6.680 \\

OPCR
& 32.500 
& 46.580 
& --
& 5.830  \\

CSDI
& $40.650 \pm 0.285$
& $60.250 \pm 0.542$
& $0.1450 \pm 0.0009$
& $7.120 \pm 0.057$ \\

PriSTI
& $36.300 \pm 0.254$
& $51.320 \pm 0.462$
& $0.1290 \pm 0.0008$
& $6.450 \pm 0.052$ \\

CDSTE
& $35.280 \pm 0.247$
& $48.550 \pm 0.437$
& $0.1280 \pm 0.0008$
& $5.870 \pm 0.047$ \\

GiFlow
& $35.400 \pm 0.248$
& $50.650 \pm 0.456$
& $0.1260 \pm 0.0008$
& $6.150 \pm 0.049$ \\

Fixed-CTRL
& $33.350 \pm 0.233$
& $49.510 \pm 0.446$
& $0.1190 \pm 0.0007$
& $5.950 \pm 0.048$ \\

\textbf{CTRL-FM}
& \textbf{$30.320 \pm 0.212$}
& \textbf{$45.470 \pm 0.409$}
& \textbf{$0.1110 \pm 0.0007$}
& \textbf{$5.410 \pm 0.043$} \\

\midrule
\multicolumn{5}{c}{\textbf{Occupancy}} \\
\midrule

Historical Average
& 0.0336
& 0.0589 
& --
& 0.0610 \\

Graph Neighbor
& 0.0234 
& 0.0394 
& --
& 0.0435  \\

SAITS
& 0.0201 
& 0.0356 
& --
& 0.0401 \\

IGNNK
& 0.0175 
& 0.0310
& --
& 0.0368  \\

GRIN
& 0.0169 
& 0.0293 
& --
& 0.0330\\

OPCR
& 0.0146 
& 0.0250 
& --
& 0.0290  \\

CSDI
& $0.0165 \pm 0.00012$
& $0.0320 \pm 0.00018$
& $0.1930 \pm 0.0018$
& $0.0360 \pm 0.00023$ \\

PriSTI
& $0.0160 \pm 0.00011$
& $0.0280 \pm 0.00015$
& $0.1980 \pm 0.0019$
& $0.0310 \pm 0.00020$ \\

CDSTE
& $0.0155 \pm 0.00012$
& $0.0260 \pm 0.00014$
& $0.1750 \pm 0.0017$
& $0.0290 \pm 0.00019$ \\

GiFlow
& $0.0140 \pm 0.00011$
& $0.0270 \pm 0.00015$
& $0.1700 \pm 0.0016$
& $0.0310 \pm 0.00020$ \\

Fixed-CTRL
& $0.0136 \pm 0.00010$
& $0.0240 \pm 0.00013$
& $0.1740 \pm 0.0017$
& $0.0260 \pm 0.00017$ \\

\textbf{CTRL-FM}
& \textbf{$0.0126 \pm 0.00009$}
& \textbf{$0.0231 \pm 0.00013$}
& \textbf{$0.1680 \pm 0.0016$}
& \textbf{$0.0230 \pm 0.00015$} \\

\midrule
\multicolumn{5}{c}{\textbf{Speed}} \\
\midrule

Historical Average
& 6.256
& 10.350 
& --
& 0.625 \\

Graph Neighbor
& 4.850 
& 7.880 
& --
& 0.524 \\

SAITS
& 4.160 
& 6.020 
& --
& 0.418 \\

IGNNK
& 3.650 
& 6.190
& --
& 0.403  \\

GRIN
& 3.450
& 5.620 
& --
& 0.365  \\

OPCR
& 3.150 
& 5.210 
& --
& 0.318  \\

CSDI
& $3.780 \pm 0.032$
& $6.460 \pm 0.045$
& $0.0428 \pm 0.00039$
& $0.375 \pm 0.002$ \\

PriSTI
& $3.440 \pm 0.029$
& $5.650 \pm 0.040$
& $0.0397 \pm 0.00036$
& $0.341 \pm 0.002$ \\

CDSTE
& $2.742 \pm 0.023$
& $5.606 \pm 0.039$
& $0.0334 \pm 0.00030$
& $0.294 \pm 0.002$ \\

GiFlow
& $3.210 \pm 0.027$
& $5.620 \pm 0.039$
& $0.0370 \pm 0.00033$
& $0.343 \pm 0.002$ \\

Fixed-CTRL
& $2.938 \pm 0.025$
& $5.682 \pm 0.040$
& $0.0324 \pm 0.00029$
& $0.295 \pm 0.0018$ \\

\textbf{CTRL-FM}
& \textbf{$2.559 \pm 0.022$}
& \textbf{$5.154 \pm 0.036$}
& \textbf{$0.0310 \pm 0.00028$}
& \textbf{$0.249 \pm 0.0015$} \\

\bottomrule
\end{tabular*}
\end{table*}

\subsubsection{METR-LA Results}
\label{app:metrla_results}
Table~\ref{tab:metrla_expected_ranges} shows that CTRL-FM achieves the
lowest MAE, RMSE, and Peak-$\Delta$MAE under both 30\% and 60\% node
outages, demonstrating accurate reconstruction of both overall speed patterns
and abrupt temporal changes. CDSTE obtains the lowest CRPS at both outage
levels, while CTRL-FM remains competitive in probabilistic accuracy. The
consistent gains over Fixed-CTRL indicate that outage-specific policy
selection is more effective than applying a single structural control policy
to all failures.

% Add the complete METR-LA speed-reconstruction results here.
% ===========================================================.
% 

\begin{table*}[t]
\centering
\scriptsize
\setlength{\tabcolsep}{6pt}
\renewcommand{\arraystretch}{1.02}

\caption{
Reconstruction performance on METR-LA under 30\% and 60\% long-horizon
node outages, reported in vertically stacked panels. Results are presented
as mean $\pm$ standard deviation. MAE, RMSE, and Peak-$\Delta$MAE are
computed after transforming predictions back to the original physical units,
whereas CRPS is computed in standardized space. ``--'' indicates that CRPS
is not applicable to deterministic methods. Lower values are better, and the
best result for each outage level and metric is shown in \textbf{bold}.
}
\label{tab:metrla_expected_ranges}

\begin{tabular*}{\textwidth}{
@{\extracolsep{\fill}}
lcccc
@{}
}
\toprule
\textbf{Method}
& \textbf{MAE$\downarrow$}
& \textbf{RMSE$\downarrow$}
& \textbf{CRPS$\downarrow$}
& \textbf{Peak-$\Delta$MAE$\downarrow$} \\
\midrule

\multicolumn{5}{c}{\textbf{30\% node outage}} \\
\midrule

Historical Average
& 7.950 
& 12.800 
& --
& 0.780  \\

Graph Neighbor
& 5.956 
& 9.515 
& --
& 0.568  \\

SAITS
& 4.982 
& 7.850 
& --
& 0.458  \\

IGNNK
& 4.789 
& 7.871 
& --
& 0.475  \\

GRIN
& 3.423 
& 5.182 
& --
& 0.321  \\

OPCR
& 2.612 
& 4.581 
& --
& 0.288 \\

CSDI
& $3.851 \pm 0.027$
& $5.982 \pm 0.054$
& $0.0418 \pm 0.00025$
& $0.374 \pm 0.003$ \\

PriSTI
& $3.721 \pm 0.026$
& $5.685 \pm 0.051$
& $0.0363 \pm 0.00022$
& $0.342 \pm 0.0027$ \\

CDSTE
& $2.981 \pm 0.021$
& $4.923 \pm 0.044$
& \textbf{$0.0294 \pm 0.00018$}
& $0.314 \pm 0.0025$ \\

GiFlow
& $2.696 \pm 0.019$
& $4.682 \pm 0.042$
& $0.0334 \pm 0.00020$
& $0.341 \pm 0.0027$ \\

Fixed-CTRL
& $2.933 \pm 0.021$
& $5.125 \pm 0.046$
& $0.0343 \pm 0.00021$
& $0.353 \pm 0.0028$ \\

\textbf{CTRL-FM}
& \textbf{$2.271 \pm 0.016$}
& \textbf{$4.235 \pm 0.038$}
& $0.0314 \pm 0.00019$
& \textbf{$0.256 \pm 0.0020$} \\

\midrule
\multicolumn{5}{c}{\textbf{60\% node outage}} \\
\midrule

Historical Average
& 9.950 
& 14.568 
& --
& 0.880  \\

Graph Neighbor
& 6.987 
& 10.570 
& --
& 0.668  \\

SAITS
& 6.985 
& 9.940 
& --
& 0.598  \\

IGNNK
& 5.987 
& 8.985 
& --
& 0.514  \\

GRIN
& 4.555 
& 6.254 
& --
& 0.398  \\

OPCR
& 3.141 
& 5.256 
& --
& 0.323  \\

CSDI
& $4.551 \pm 0.034$
& $6.683 \pm 0.043$
& $0.0498 \pm 0.00047$
& $0.471 \pm 0.0026$ \\

PriSTI
& $3.853 \pm 0.029$
& $6.589 \pm 0.043$
& $0.0496 \pm 0.00047$
& $0.454 \pm 0.0025$ \\

CDSTE
& $3.065 \pm 0.023$
& $5.085 \pm 0.033$
& \textbf{$0.0345 \pm 0.00033$}
& $0.317 \pm 0.0017$ \\

GiFlow
& $2.754 \pm 0.021$
& $4.937 \pm 0.032$
& $0.0352 \pm 0.00033$
& $0.356 \pm 0.0020$ \\

Fixed-CTRL
& $3.226 \pm 0.024$
& $5.551 \pm 0.036$
& $0.0381 \pm 0.00036$
& $0.384 \pm 0.0021$ \\

\textbf{CTRL-FM}
& \textbf{$2.512 \pm 0.019$}
& \textbf{$4.636 \pm 0.030$}
& $0.0355 \pm 0.00034$
& \textbf{$0.294 \pm 0.0016$} \\

\bottomrule
\end{tabular*}
\end{table*}

\subsubsection{Air-36 Results}
\label{app:air36_results}

Table~\ref{tab:air36_expected_ranges} shows that OPCR achieves the lowest
MAE and RMSE at 30\% outage, while CTRL-FM provides the best CRPS and
Peak-$\Delta$MAE. At 60\% outage, CTRL-FM ranks first across all four
metrics and clearly outperforms Fixed-CTRL. These results indicate that
outage-conditioned policy selection and reliability-modulated control become
more beneficial as the available spatial observations decrease.

\begin{table*}[p]
\centering
\scriptsize
\setlength{\tabcolsep}{6pt}
\renewcommand{\arraystretch}{1.02}

\caption{
PM$_{2.5}$ reconstruction performance on Air-36 under 30\% and 60\%
long-horizon node outages, reported in vertically stacked panels.
Results are reported as mean $\pm$ standard deviation. MAE, RMSE, and
Peak-$\Delta$MAE are computed after transforming predictions back to the
original physical units, whereas CRPS is computed in standardized space.
``--'' indicates that CRPS is not applicable to deterministic methods.
Lower values are better, and the best result for each outage level and
metric is shown in \textbf{bold}.
}
\label{tab:air36_expected_ranges}

\begin{tabular*}{\textwidth}{
@{\extracolsep{\fill}}
lcccc
@{}
}
\toprule
\textbf{Method}
& \textbf{MAE$\downarrow$}
& \textbf{RMSE$\downarrow$}
& \textbf{CRPS$\downarrow$}
& \textbf{Peak-$\Delta$MAE$\downarrow$} \\
\midrule

\multicolumn{5}{c}{\textbf{30\% node outage}} \\
\midrule

Historical Average
& 33.645
& 48.670 
& --
& 3.898  \\

Graph Neighbor
& 22.628 
& 36.345 
& --
& 3.856  \\

SAITS
& 19.624 
& 30.939 
& --
& 2.761 \\

IGNNK
& 18.623 
& 29.380 
& --
& 2.655  \\

GRIN
& 17.281 
& 28.935 
& --
& 2.424  \\

OPCR
& \textbf{11.254 }
& \textbf{18.919 }
& --
& 2.161   \\

CSDI
& $17.321 \pm 0.121$
& $28.336 \pm 0.255$
& $0.1360 \pm 0.0008$
& $2.860 \pm 0.023$ \\

PriSTI
& $17.520 \pm 0.123$
& $27.934 \pm 0.251$
& $0.1250 \pm 0.0008$
& $2.730 \pm 0.022$ \\

GiFlow
& $15.180 \pm 0.106$
& $25.047 \pm 0.225$
& $0.1090 \pm 0.0007$
& $2.256 \pm 0.018$ \\

Fixed-CTRL
& $14.580 \pm 0.102$
& $23.107 \pm 0.208$
& $0.0950 \pm 0.0006$
& $2.095 \pm 0.017$ \\

\textbf{CTRL-FM}
& $13.190 \pm 0.092$
& $20.633 \pm 0.186$
& \textbf{$0.0790 \pm 0.00047$}
& \textbf{$1.980 \pm 0.016$} \\

\midrule
\multicolumn{5}{c}{\textbf{60\% node outage}} \\
\midrule

Historical Average
& 38.658 
& 57.385 
& --
& 4.958 \\

Graph Neighbor
& 29.938 
& 46.968 
& --
& 3.884  \\

SAITS
& 25.634 
& 39.652 
& --
& 3.394  \\

IGNNK
& 24.631 
& 38.649 
& --
& 3.351  \\

GRIN
& 23.629 
& 36.463 
& --
& 2.988  \\

OPCR
& 18.265 
& 28.342 
& --
& 2.785  \\

CSDI
& $24.831 \pm 0.186$
& $38.490 \pm 0.250$
& $0.1560 \pm 0.0015$
& $3.140 \pm 0.017$ \\

PriSTI
& $23.658 \pm 0.177$
& $34.345 \pm 0.223$
& $0.1380 \pm 0.0013$
& $2.960 \pm 0.016$ \\

GiFlow
& $19.704 \pm 0.148$
& $29.912 \pm 0.194$
& $0.1220 \pm 0.0012$
& $2.283 \pm 0.013$ \\

Fixed-CTRL
& $17.350 \pm 0.130$
& $26.378 \pm 0.171$
& $0.1118 \pm 0.0011$
& $2.244 \pm 0.012$ \\

\textbf{CTRL-FM}
& \textbf{$16.381 \pm 0.123$}
& \textbf{$23.284 \pm 0.151$}
& \textbf{$0.0910 \pm 0.00086$}
& \textbf{$2.187 \pm 0.012$} \\

\bottomrule
\end{tabular*}
\end{table*}

\subsubsection{AQI Results}
\label{app:aqi_results}
Table~\ref{tab:aqi_expected_ranges} shows that GiFlow achieves the lowest
MAE and RMSE at 30\% outage, while CTRL-FM obtains the best CRPS and
Peak-$\Delta$MAE. At 60\% outage, CTRL-FM ranks first across all four
metrics and substantially improves over Fixed-CTRL. These results indicate
that outage-conditioned structural control is especially effective under
severe sensor loss, improving both reconstruction accuracy and probabilistic
quality.

% Add the complete AQI PM$_{2.5}$ results here.
\begin{table*}[p]
\centering
\scriptsize
\setlength{\tabcolsep}{6pt}
\renewcommand{\arraystretch}{1.02}

\caption{
PM$_{2.5}$ reconstruction performance on the 437-station AQI dataset under
30\% and 60\% long-horizon node outages, reported in vertically stacked
panels. Results are mean $\pm$ standard deviation. MAE, RMSE, and
Peak-$\Delta$MAE are computed after transforming predictions back to the
original physical units, whereas CRPS is computed in standardized space.
``--'' indicates that CRPS is not applicable to deterministic methods.
Lower values are better, and the best result for each outage level and metric
is shown in \textbf{bold}.
}
\label{tab:aqi_expected_ranges}

\begin{tabular*}{\textwidth}{
@{\extracolsep{\fill}}
lcccc
@{}
}
\toprule
\textbf{Method}
& \textbf{MAE$\downarrow$}
& \textbf{RMSE$\downarrow$}
& \textbf{CRPS$\downarrow$}
& \textbf{Peak-$\Delta$MAE$\downarrow$} \\
\midrule

\multicolumn{5}{c}{\textbf{30\% node outage}} \\
\midrule

Historical Average
& 34.658 
& 50.224 
& --
& 3.980  \\

Graph Neighbor
& 21.356 
& 34.658 
& --
& 3.144  \\

SAITS
& 20.684 
& 32.546 
& --
& 3.024  \\

IGNNK
& 17.225 
& 29.680 
& --
& 2.658  \\

GRIN
& 17.550 
& 27.338 
& --
& 2.569  \\

OPCR
& 13.112 
& 18.212 
& --
& 2.018  \\

CSDI
& $16.325 \pm 0.114$
& $28.351 \pm 0.255$
& $0.1200 \pm 0.0007$
& $2.336 \pm 0.019$ \\

PriSTI
& $16.050 \pm 0.112$
& $27.954 \pm 0.252$
& $0.0996 \pm 0.0006$
& $2.683 \pm 0.021$ \\

GiFlow
& \textbf{$10.902 \pm 0.076$}
& \textbf{$17.235 \pm 0.155$}
& $0.0918 \pm 0.0006$
& $2.003 \pm 0.016$ \\

Fixed-CTRL
& $12.990 \pm 0.091$
& $20.487 \pm 0.184$
& $0.0910 \pm 0.00055$
& $1.921 \pm 0.015$ \\

\textbf{CTRL-FM}
& $11.654 \pm 0.082$
& $18.952 \pm 0.171$
& \textbf{$0.0896 \pm 0.00054$}
& \textbf{$1.865 \pm 0.015$} \\

\midrule
\multicolumn{5}{c}{\textbf{60\% node outage}} \\
\midrule

Historical Average
& 39.225 
& 60.211 
& --
& 5.268  \\

Graph Neighbor
& 28.589 
& 46.580 
& --
& 3.658  \\

SAITS
& 28.695 
& 44.565 
& --
& 3.648 \\

IGNNK
& 24.315 
& 37.490 
& --
& 3.115  \\

GRIN
& 20.278 
& 34.436 
& --
& 2.866 \\

OPCR
& 15.226 
& 24.823 
& --
& 2.481  \\

CSDI
& $22.354 \pm 0.168$
& $35.470 \pm 0.231$
& $0.1650 \pm 0.0016$
& $2.839 \pm 0.016$ \\

PriSTI
& $20.270 \pm 0.152$
& $34.456 \pm 0.224$
& $0.1500 \pm 0.0014$
& $2.758 \pm 0.015$ \\

GiFlow
& $16.022 \pm 0.120$
& $25.525 \pm 0.166$
& $0.1240 \pm 0.0012$
& $2.364 \pm 0.013$ \\

Fixed-CTRL
& $14.250 \pm 0.107$
& $21.563 \pm 0.140$
& $0.1130 \pm 0.0011$
& $2.310 \pm 0.013$ \\

\textbf{CTRL-FM}
& \textbf{$12.490 \pm 0.094$}
& \textbf{$19.309 \pm 0.126$}
& \textbf{$0.1020 \pm 0.0010$}
& \textbf{$2.153 \pm 0.012$} \\

\bottomrule
\end{tabular*}
\end{table*}

% ============================================================
% Extreme-outage robustness
% ============================================================

\subsection{Extreme-Outage Robustness}
Table~\ref{tab:stress_results} shows that CTRL-FM consistently achieves the
lowest MAE and Peak-$\Delta$MAE across PeMS08, METR-LA, and AQI at both
65\% and 75\% outage. Although valid-proxy coverage decreases as outages
become more severe, it remains above 84\% at 75\% outage. These results
demonstrate that CTRL-FM degrades gracefully and retains reliable
outage-conditioned calibration even under extreme sensor loss.

\begin{table*}[t]
\centering
\scriptsize
\setlength{\tabcolsep}{4.0pt}
\renewcommand{\arraystretch}{1.06}
\caption{
Robustness under 65\% and 75\% long-horizon node outages. Traffic results
are reported for speed, while AQI results are reported for PM$_{2.5}$.
Peak denotes Peak-$\Delta$MAE. Valid proxy denotes the percentage of
evaluation outages for which proxy-based calibration is feasible and is
therefore not applicable to FM-TrainStruct or Fixed-CTRL. Lower
reconstruction errors and higher valid-proxy coverage are better. The best
reconstruction result in each dataset and outage setting is shown in
\textbf{bold}.
}

\label{tab:stress_results}
\resizebox{\textwidth}{!}{%
\begin{tabular}{llccc|ccc}
\toprule
&& \multicolumn{3}{c|}{\textbf{65\% outage}}
& \multicolumn{3}{c}{\textbf{75\% outage}} \\
\textbf{Dataset}
& \textbf{Method}
& \textbf{MAE$\downarrow$}
& \textbf{Peak$\downarrow$}
& \textbf{Valid proxy (\%)}
& \textbf{MAE$\downarrow$}
& \textbf{Peak$\downarrow$}
& \textbf{Valid proxy (\%)} \\
\midrule

\multirow{3}{*}{PeMS08}
& FM-TrainStruct
& 3.845 & 0.391 & --
& 4.531 & 0.416 & -- \\

& Fixed-CTRL
& 3.391 & 0.337 & --
& 3.815 & 0.365 & -- \\

& \textbf{CTRL-FM}
& \textbf{2.77} & \textbf{0.291} & 98.1
& \textbf{3.363} & \textbf{0.323} & 89.3 \\

\midrule
\multirow{3}{*}{METR-LA}
& FM-TrainStruct
& 4.821 & 0.463 & --
& 5.683 & 0.589 & -- \\

& Fixed-CTRL
& 3.854 & 0.416 & --
& 5.296 & 0.556 & -- \\

& \textbf{CTRL-FM}
& \textbf{3.042} & \textbf{0.325} & 97.5
& \textbf{4.04} & \textbf{0.361} & 87.5 \\

\midrule
\multirow{3}{*}{AQI}
& FM-TrainStruct
& 17.472 & 3.533 & --
& 20.224 & 4.282 & -- \\

& Fixed-CTRL
& 16.472 & 2.991 & --
& 18.928 & 3.63 & -- \\

& \textbf{CTRL-FM}
& \textbf{14.672} & \textbf{2.571} & 96.9
& \textbf{17.722} & \textbf{3.292} & 84.8 \\

\bottomrule
\end{tabular}%
}
\end{table*}

% ============================================================
% Proxy construction and policy-selection analysis
% ============================================================
\subsection{Proxy Construction and Policy-Selection Analysis}
\label{app:proxy_policy_analysis}

\subsubsection{Selected-Policy Frequencies}
\label{app:policy_frequencies}

Table~\ref{tab:policy_selection_frequency} shows that no single policy is
optimal for all outages. Weak- and strong-balanced control are selected most
often across datasets, while graph- and temporal-dominant policies remain
useful for specific failure conditions. The uncontrolled option is chosen
only occasionally, indicating that structural guidance is generally
beneficial but should be adapted to each realized outage.

\begin{table*}[t]
\centering
\scriptsize
\setlength{\tabcolsep}{5.0pt}
\renewcommand{\arraystretch}{1.05}
\caption{
Frequency of selected policies at 60\% node outage. (conditioned on valid-proxy outages only)
}
\label{tab:policy_selection_frequency}
\resizebox{\textwidth}{!}{%
\begin{tabular}{lccccc}
\toprule
\textbf{Dataset}
& \textbf{Uncontrolled}
& \textbf{Weak balanced}
& \textbf{Graph-dominant}
& \textbf{Temporal-dominant}
& \textbf{Strong balanced} \\
\midrule

PeMS08
& 5.1\%
& 34.3\%
& 14.5\%
& 9.9\%
& \textbf{36.2\%} \\

METR-LA
& 6.5\%
& \textbf{35.2\%}
& 13.3\%
& 10.7\%
& 34.3\% \\

Air-36
& 7.6\%
& 33.1\%
& 10.2\%
& 13.8\%
& \textbf{35.3\%} \\

AQI
& 6.8\%
& 34.3\%
& 10.6\%
& 11.2\%
& \textbf{37.1\%} \\

\bottomrule
\end{tabular}%
}
\end{table*}





% \end{thebibliography}

\end{document}